%% file: main.tex
\documentclass[twocolumn,letterpaper]{IEEEAerospaceCLS}  %

\usepackage{siunitx}

\usepackage[]{graphicx}    %
\usepackage{amsmath}
\usepackage{amssymb}
\usepackage[caption=false]{subfig}
\newcommand{\ignore}[1]{}  %

\begin{document}
\title{Machine Vision based Sample-Tube Localization for Mars Sample Return}

\author{%
Shreyansh Daftry, Barry Ridge, William Seto, Tu-Hoa Pham, Peter Ilhardt$^{*}$, Gerard Maggiolino$^{*}$, \\
Mark Van der Merwe$^{*}$, Alex Brinkman, John Mayo, Eric Kulczyski and Renaud Detry\\ 
Jet Propulsion Laboratory, California Institute of Technology\\
Pasadena, CA, USA\\
\texttt{Shreyansh.Daftry@jpl.nasa.gov}\\
\thanks{$^{*}$P. Ilhardt, G. Maggiolino and M. Merwe were interns at JPL during this work, and are currently affiliated with Capgemini, Carnegie Mellon University and University of Michigan, respectively.}\\
\thanks{\footnotesize 978-1-7281-7436-5/21/$\$31.00$ \copyright2021 IEEE} %
}

\maketitle

\thispagestyle{plain}
\pagestyle{plain}

\maketitle

\thispagestyle{plain}
\pagestyle{plain}

\begin{abstract} 
  A potential Mars Sample Return (MSR) architecture is being jointly studied by NASA and ESA. As currently envisioned, the MSR campaign consists of a series of 3 missions: sample cache, fetch and return to Earth. In this paper, we focus on the fetch part of the MSR, and more specifically the problem of autonomously detecting and localizing sample tubes deposited on the Martian surface. Towards this end, we study two machine-vision based approaches: First, a geometry-driven approach based on template matching that uses hard-coded filters and a 3D shape model of the tube; and second, a data-driven approach based on convolutional neural networks (CNNs) and learned features. Furthermore, we present a large benchmark dataset of sample-tube images, collected in representative outdoor environments and annotated with ground truth segmentation masks and locations. The dataset was acquired systematically across different terrain, illumination conditions and dust-coverage; and benchmarking was performed to study the feasibility of each approach, their relative strengths and weaknesses, and robustness in the presence of adverse environmental conditions.
\end{abstract} 

\setcounter{tocdepth}{1}
\tableofcontents

\section{Introduction}
\input{sec_intro}
\label{sec:intro}

\section{Related Work}
\input{sec_related}
\label{sec:related}

\section{Sample-tube Localization}

\input{sec_overview}
\label{sec:overview}

\section{Template-based Object Detection}
\input{sec_template_matching}

\label{sec:tm}

\section{Data-driven Segmentation}
\input{sec_maskrcnn}
\label{sec:maskrcnn}

\section{Benchmarking Dataset}
\input{sec_dataset}
\label{sec:dataset}

\section{Experiments and Results}
\input{sec_experiment}

\label{sec:exp}

\section{Conclusion and Future Work}
\input{sec_conclusion}
\label{sec:conclusion}

\acknowledgments
The research described in this paper was carried out at the Jet Propulsion Laboratory, California Institute of Technology, under a contract with the National Aeronautics and Space Administration.

\bibliographystyle{IEEEtran}
\bibliography{reference,renaud}

\thebiography
\input{sec_biography}

\end{document}

%% file: sec_intro.tex
Determining the habitability of past and present martian environments continues to be the focus of current and future missions to Mars. Indeed, studying the geological history of Mars holds the key to both understanding the origins of life on Earth and in the Solar System. While recent and ongoing robotic missions have revolutionized our understanding of the red planet \cite{squyres2006rocks,grotzinger2013analysis}, the results from orbital and in-situ surface robotic missions alone are not sufficient to fully answer the major questions about the potential for life, past climate, and the geological history of Mars. Even if an orbital or in situ mission were to discover putative evidence for the existence of past or present life on Mars, confirming these results would necessitate that samples be collected, returned to Earth, and verified by multiple rigorous laboratory analyses on Earth. As a result, returning samples from Mars back to Earth was identified as the highest priority planetary science objective in the Planetary Science Decadal Survey \cite{board2012vision}.

\begin{figure}[!t]
    \centering
    \includegraphics[width=\linewidth]{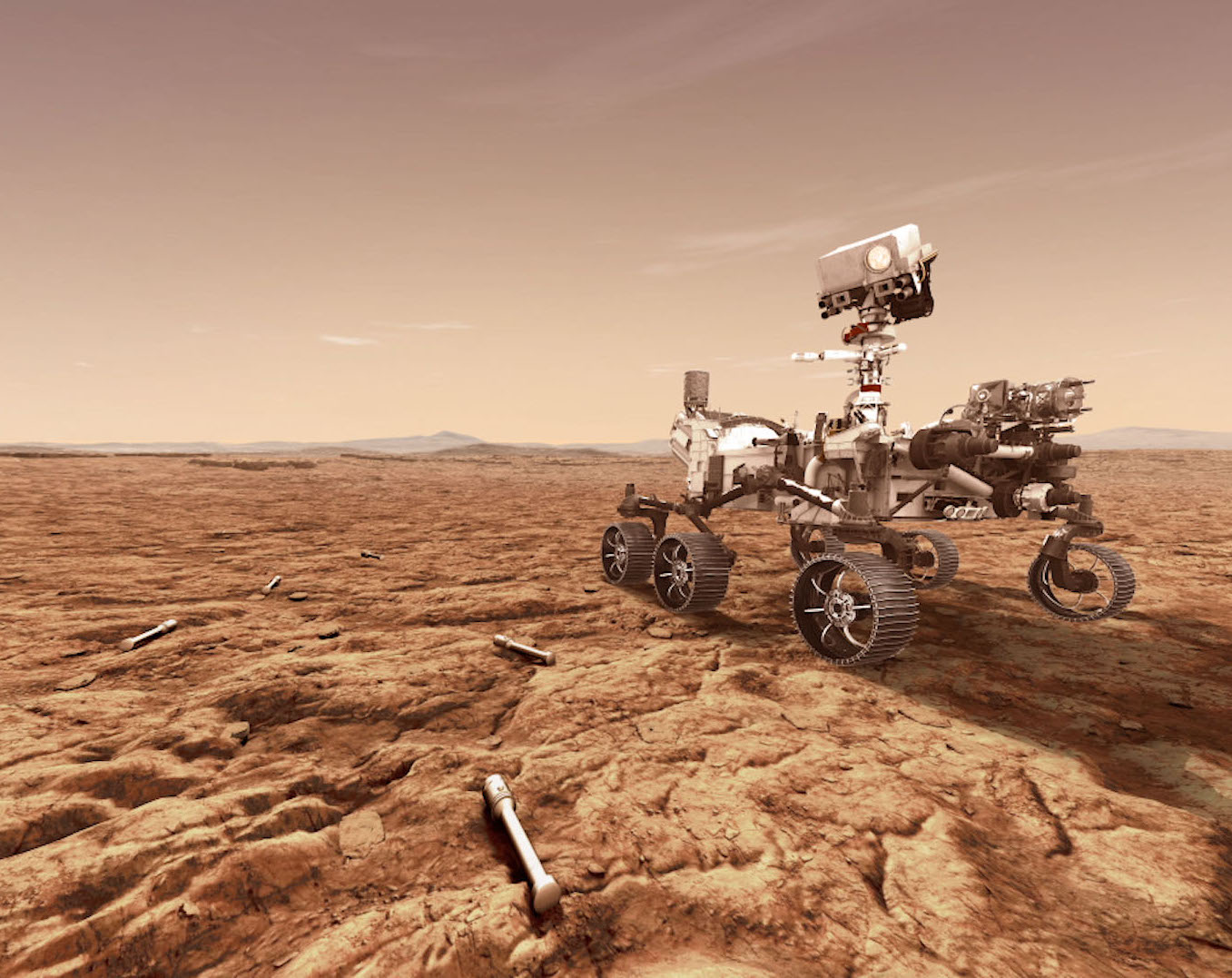}
    \caption{NASA's Mars Perseverance rover, which serves as the first stage of the MSR campaign, will store rock and soil samples in sealed tubes on the planet's surface for future missions to retrieve, as seen in this illustration.}
    \label{fig:teaser}
\end{figure}

In response to that, a potential concept for a Mars Sample Return (MSR) architecture is being jointly studied by NASA and ESA \cite{muirhead2020mars}. As currently envisioned, the MSR campaign consists of a series of 3 missions: sample cache, fetch and return to Earth. First, NASA’s Perseverance Rover, launched in 2020, will collect scientifically selected samples and store them in sealed tubes on the planet’s surface (see Figure\ref{fig:teaser}), for possible return to Earth. Then, a potential future mission, with a Sample Retrieval Lander (SRL), would collect the sample tubes and load them into an Orbiting Sample (OS) payload in a Mars Ascent Vehicle (MAV). The MAV would release the OS into Martian orbit. The third mission, an Earth Return Orbiter (ERO), would rendezvous with the samples in Mars orbit and ferry them back to Earth.

This work focuses on the SRL mission concept \cite{muirhead2019sample} to collect and retrieve the sample tubes. The SRL mission would deploy a lander in the vicinity of Jezero Crater, where the Mars Perseverance rover plans to land and collect and cache samples during its 1.25-Mars-year primary surface mission. Key payloads on SRL would include an ESA-provided Sample Fetch Rover (SFR) and Sample Transfer Arm (STA), and a NASA-provided OS and MAV. Once on the surface, the SFR would egress from SRL and begin its surface mission to retrieve samples previously cached by Mars Perseverance Rover at one or more depot locations. The solar-powered SRL mission would then carry out its surface activities during Martian spring and summer, maximizing available power, and would complete its surface mission and launch the retrieved samples into orbit before start of northern hemisphere fall, prior to significant decrease in available solar power and in advance of the potential for global dust storms. According to the currently envisioned surface mission timeline, this would allocate $\sim150$ sols for SFR to complete this retrieval and return to SRL. This constrained surface mission timeline, combined with SFR drive distances predicted to be up to 4-km roundtrip, drives the need for high levels of SFR autonomy to enable efficient tube pickup and transfer operations.

In this paper, we study the problem of autonomously detecting and localizing sample tubes deposited on the Martian surface directly from camera images. In Section \ref{sec:tm} we discuss the first of two machine-vision approaches that we studied, a geometry-driven approach based on template matching that uses hard-coded filters and a 3D shape model of the tube; and in Section \ref{sec:maskrcnn} we discuss the second, a data-driven approach based on convolutional neural networks (CNNs) and learned features. In Section \ref{sec:dataset}, we present a large benchmark dataset of sample-tube images, collected in representative outdoor environments and annotated with ground truth segmentation masks and locations. The dataset was acquired systematically across different terrain, illumination conditions and dust-coverage. Finally, in Section \ref{sec:exp}, we describe the benchmarking evaluations that were performed using well-known metrics to compare and assess the two methods using this data.

%% file: sec_related.tex
\subsection{Mars Sample Return}
Several studies have been conducted over the last few decades with the goal of defining science objectives for Mars Sample Return \cite{board2012vision,mepag2008science,mclennan2012planning,beaty2019a} and potential mission concept architectures \cite{sherwood2002mars,mattingly2004continuing,mattingly2011a,muirhead2020a}. Furthermore, to enable this vision, several research and technology development efforts \cite{volpe2000technology} have been under way - from autonomous rover technology \cite{weisbin1999autonomous,osinski2019canmars} to OS design and capture-systems \cite{younse2020concept,perino2017evolution}. Our work is closely related to, and a continuation of, the efforts to demonstrate robust localization and autonomous retrieval of sample-tubes from a Mars-like environment for potential Mars Sample Return. While \cite{edelberg2015autonomous} served as an initial proof-of-concept, subsequent work \cite{papon2017a,lee2018monocular,pham2020rover} further studied the problem of direct and indirect sample-tube localization in an analog indoor testbed. In this work, we do a comprehensive study on the performance of sample-tube detection algorithms on a benchmark dataset collected in a representative outdoor environment.

\subsection{Image-based Object Detection and Localization}
Object detection and localization, as of one the most fundamental and challenging problems in machine vision, has received great attention in recent years - see \cite{zou2019object} for a review. Traditionally, the problem of object detection and localization is tackled by matching feature points between 3D models and images \cite{lowe1999object,rothganger20063d}. However, these methods require that there are rich textures on the objects in order to detect features for matching. As a result they are unable to handle texture-less objects like sample-tubes. More recently, with the advent of deep learning, CNNs by comparison, have made significant progress in object classification \cite{krizhevsky2012imagenet}, detection \cite{girshick2015fast,ren2015faster}, semantic segmentation \cite{long2015fully} and instance segmentation \cite{he2017mask}, including their application to Mars rover autonomy \cite{ono2020maars,abcouwer2020machine}.

For texture-less object detection, the taxonomy of approaches can be broadly classified into three categories: feature-based, view-based and shape-based. The feature-based approaches \cite{costa20003d,david2005object,weiss2001model} match 2D image features like edges or line segments to the corresponding 3D model points. The view-based approaches \cite{cyr2004similarity,eggert1993scale,ulrich2011combining} compare a given image of the object with its pre-computed 2D views. Lastly, shape-based approaches \cite{hinterstoisser2011gradient,hinterstoisser2012model,cai2013fast,tombari2010a,tsai2018real} are based on template matching of edge-segments. Furthermore, end-to-end learning based methods \cite{do2018deep,kehl2017ssd,xiang2017posecnn} have also been proposed that utilize deep learning for the task of 6D pose estimation directly from RGB images as the only input modality.

%% file: sec_overview.tex
The operational context assumed is as follows. The Perseverance rover will place sample tubes on the ground at one or multiple (presently-unknown) sites referred to as \emph{sample depots}. Within each sample depot, the sample-tubes will be placed several meters apart in areas where sand is not abundant and that have a relatively flat ground with slopes less than \SI{10}{\degree}. It is to be noted that Perseverance has little control over precise tube placement: tubes will be released from the rover's underbelly, and may bounce and roll on the ground before coming to a halt. Furthermore, it is anticipated that SFR is capable of autonomously driving from tube to tube -- a capability addressed in a separate paper \cite{pham2021rover}. For the context of this paper, this implies that SFR can drive itself to a pose that places a tube within its field of view.

Based on these assumptions, our understanding of the Martian conditions (weather, geology, etc.) and through qualitative observation of past lander and rover images, we can conservatively hypothesize the following constraints: (a) tubes will not move, (b) sand or dust may, with low probability, pile up next to tube or rocks, forming drifts, and (c) dust can deposit everywhere, potentially creating a dust layer that will not exceed 0.25mm in thickness. Rovers typically operate between 10am and 4pm, yielding a large range of lighting conditions, including shadows incurred by rocks or by the rover itself.

Our goal is to then to study the robustness for image-based detection and localization of sample-tubes under these constraints. More specifically, we quantify the performance of two object localizers for sample-tube detection in a variety of of conditions that include clear, non-occluded tubes, partial occlusions by rocks or sand, or partial or complete shadows. We focus on the detection and localization only from a single images for two reasons: first, it allows the proposed algorithms to be applicable to a diverse set of scenarios (e.g. if we want to detect using either a mast-mounted stereo camera or a wrist-mounted monocular camera), and second, based on our preliminary analysis we expect depth data on the sample-tubes to be noisy and thereby not very helpful for the detection task. Extensive analysis on the usefulness of stereo range data is beyond the scope of this paper.

The two localizers considered in this paper are a template-based object detector \cite{hinterstoisser2011gradient}, and a region-proposal network \cite{he2017mask}. Template-based object detection has more industrial maturity, and it shares building blocks with methods that ran on spacecrafts \cite{francis2017a,kim2005a}. Template matching lends itself to introspection and ad-hoc treatment of edge cases. By contrast, convolutional networks have almost no flight heritage, and their design is harder to validate, but their performance is often substantially better than template matching. 

%% file: sec_template_matching.tex
\begin{figure}[!t]
    \centering
    \includegraphics[width=0.95\linewidth]{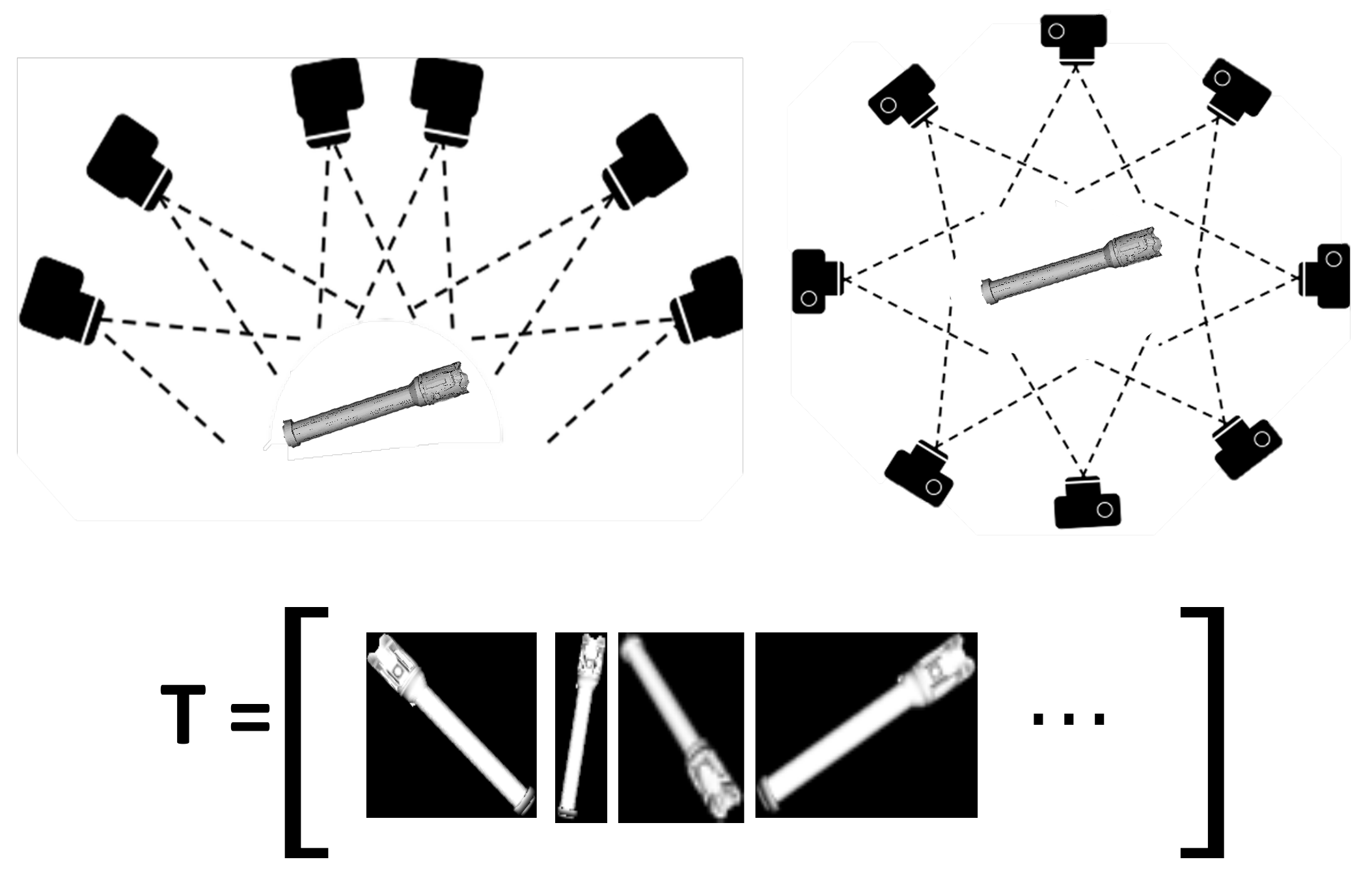}
    \caption{Sample set of templates $\mathcal{T}$ generated using different viewpoints sampled around the sample-tube.}
    \label{fig:template-generauin}
\end{figure}

Our approach is based on Line2D \cite{hinterstoisser2011gradient,hinterstoisser2012model}, an efficient template-matching based method that exploits color images to capture the appearance of the object in a set of templates covering different views. Because the viewpoint of each template is known, it also provides a coarse estimate of the 6D pose of the object when it is detected. In the remainder of this section, we give an overview of the Line2D method and how we use it to detect and localize sample-tubes.

\subsection{Generating Templates using a 3D Model}
Given a 3D CAD model of the sample-tube, a library of 2D templates are automatically generated that covers a full view hemisphere by regularly sampling viewpoints of the 3D model, as illustrated in Figure \ref{fig:template-generauin}. During the generation of template library, it is important to balance the trade-off between the coverage of the object for reliability and the number of templates for efficiency. This is solved by recursively dividing an icosahedron, the largest convex regular polyhedron. The vertices of the resulting polyhedron give us the two out-of-plane rotation angles for the samples pose with respect to the coordinate center. In our experiments, two adjacent vertices are set to be approximately $10$ degrees apart. In addition to the these two out of plane rotations, templates for different in-plane rotations are also created. Furthermore, templates at different scales are generated by using different sized polyhedrons, using a step size of $10$ cm.

\subsection{Gradient Orientation Features}
For each sampled pose generated by the method described above, the sample-tube's silhouette is first computed by projecting its 3D model under this pose. The silhouette contour is then quickly obtained by subtracting the eroded silhouette from its original version. Next, all the color gradients that lie on the silhouette contour are computed. Image gradients are chosen as features because they have been proved to be more discriminant than other forms of representation \cite{lowe2004distinctive}. Additionally, image gradients are often the only reliable image cue when it comes to texture-less objects. Furthermore, considering only the orientation of the gradients and not their norms makes the measure robust to contrast changes, and taking the absolute value of cosine between them allows it to correctly handle object occluding boundaries: It will not be affected if the object is over a dark background, or a bright background. 

\subsection{Template Matching}
Template matching is done by measuring the similarity between an input image $\mathcal{I}$, and a reference image $\mathcal{O}$ of the sample-tube centered on a location $c$ in the image  $\mathcal{I}$ by comparing the gradient orientation features. A model or template $\mathcal{T}$ is defined as a pair $\mathcal{T}$ = ($\mathcal{O}$, $\mathcal{P}$), where $\mathcal{P}$ specifies a region in $\mathcal{O}$. The template can then be compared with a region at location $c$ in a test image $\mathcal{I}$ based on a modified version of the similarity measure proposed by Steger \cite{steger2002occlusion}:\\[6pt]
\begin{equation} \label{eq:1}
    \varepsilon = \sum_{ \text{r} \in \mathcal{P}} \Big(\max_{ \text{t} \in \mathcal{R}( \text{c} + \text{r})}\big|\cos\big(\text{ori}(\mathcal{O}, \text{r})-\text{ori}(\mathcal{I}, \text{t})\big)\big|\Big) \\
\end{equation}\\[6pt]
where ori($\mathcal{O}$, r) is the gradient orientation at location $r$ and $\mathcal{R}$($c$ + $r$) defines the neighbourhood of size $T$ centered on location $c$ + $r$.
 
In order to avoid evaluating the $\max$ operator in Equation \ref{eq:1} every time a new template must be evaluated against an image location, a binary representation of the feature space is used \cite{rios2013discriminatively}. First, the gradient orientation map is quantized by dividing it into $n_{0}$ equal spacing. To make the quantization robust to noise, we assign to each location the gradient whose quantized orientation occurs most often in a $3$x$3$ neighborhood. Next, the possible combination of orientation spreads to a given location $m$ is encoded using a binary string. These strings are then used as indices to access lookup tables for fast pre-computation of the similarity measure. Since the lookup tables are computed offline and shared between the templates, matching several templates against the input image can be done very fast once the maps are computed.

%% file: sec_maskrcnn.tex
\subsection{Region Proposal CNNs}
Modern CNNs are powerful models that leverage large labeled datasets in order to automatically derive visual feature hierarchies directly from the data such that a learning task, e.g. object classification, can be solved. In this section we describe our use of the popular CNN-based object instance segmentation model known as Mask R-CNN \cite{he2017mask} for tube localization. The ``R'' in ``R-CNN'' stands for ``region-based'' and refers to a class of CNN networks that typically extend the classification capabilities of their forebears (e.g. \cite{he2016deep,xie2017aggregated,lin2017feature}) by adding a regression head to the architecture such that they can predict continuous values for object region proposal as well as object class labels. This can be used to predict bounding box parameters of the objects under consideration, for example, such that the objects may be localized within images as opposed to solely having their presence detected \cite{girshick2014rich}.

\subsection{Instance Segmentation with Mask R-CNN}
Mask R-CNN builds on its predecessors Fast R-CNN \cite{girshick2015fast} and Faster R-CNN \cite{ren2015faster}, which were limited to bounding box prediction, by additionally allowing for segmentation masks to be predicted. While related models like the Fully Convolutional Network (FCN) \cite{long2015fully} have dealt with \emph{semantic segmentation}, that is segmenting an image and associating the segments to the object classes to which they belong, Mask R-CNN tackles the more challenging problem of \emph{instance segmentation} whereby individual instances of objects from those classes must be identified and segmented. For our purposes in solving the task of sample tube localization, all of these methods are viable candidates, but the instance segmentation provided by Mask-RCNN enables candidate tube instances to be directly identified and segmented without needing to further process the bounding box or segmentation results provided by the other methods.

Mask-RCNN uses the architecture of Faster R-CNN to both predict the class label and to regress the bounding box parameters, but it augments this with an FCN-based mask prediction branch in order to additionally predict the regional segmentation masks in parallel. There is some generality to the architecture in that a number of possible CNN backbones can be used for the convolutional feature detection component of the Faster R-CNN branch, e.g. ResNet \cite{he2016deep}, ResNeXt \cite{xie2017aggregated} or Feature Pyramid Networks (FPN) \cite{lin2017feature}. In our experiments for this work we used ResNet-50 exclusively \cite{he2016deep}. For training, Mask R-CNN makes use of a multi-task loss $L = L_\textit{cls} + L_\textit{box} + L_\textit{mask}$ that is comprised of the classification loss $L_\textit{cls}$ and the bounding-box loss $L_\textit{box}$ from \cite{girshick2015fast}, as well as a segmentation mask loss $L_\textit{mask}$. $L_\textit{mask}$ is defined by taking the average binary cross-entropy loss over the per-pixel sigmoid-activated output of the mask prediction branch which is made up of $K$ $m \times m$ resolution output masks for each predicted region of interest, where $K$ is the number of ground-truth classes.

\subsection{Using Mask R-CNN for Tube Localization} 
As is typical, the Mask R-CNN model we employ uses model weights pre-trained on the ImageNet dataset \cite{krizhevsky2012imagenet,russakovsky2015imagenet} and transfer learning is used to fine-tune the weights for our particular task. The tube localization problem presented here is relatively simple in terms of the classification component since we need only detect the presence or absence of tubes in the scene, so we make use of only a single \emph{``tube''} class label alongside the background class. With regard to localizing the tube, given the fetch rover grasping requirements, as well as for a direct comparison with the template matching method described in the previous section, we are less interested in the bounding box output than we are in obtaining the segmentation masks. Thus, although we annotate the training data with each of the class, bounding box and segmentation mask ground truth labels and train our Mask R-CNN model with all three of them, for the purposes of this paper, at inference time we discard the bounding box output and we restrict our subsequent experimental analyses to the segmentation masks.

%% file: sec_dataset.tex
\begin{figure}[!t]
    \centering
    \subfloat[Camera setup at the Mars Yard at JPL.]{
        \includegraphics[height=0.45\columnwidth]{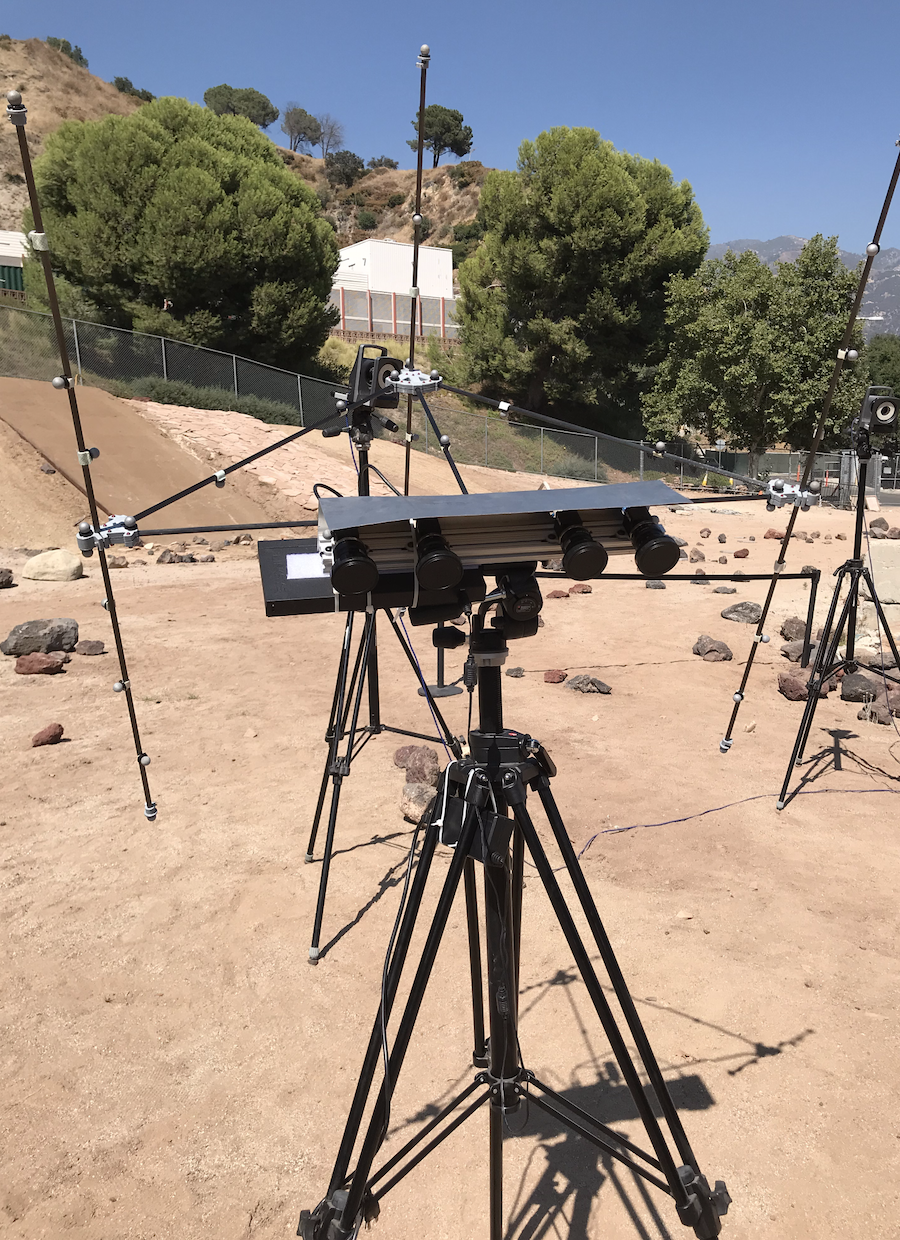}
        \label{fig:outdoordataset:camera_setup}
    } \hspace{0.1\columnwidth}
    \subfloat[Camera positioning around tubes.]{
        \includegraphics[height=0.45\columnwidth]{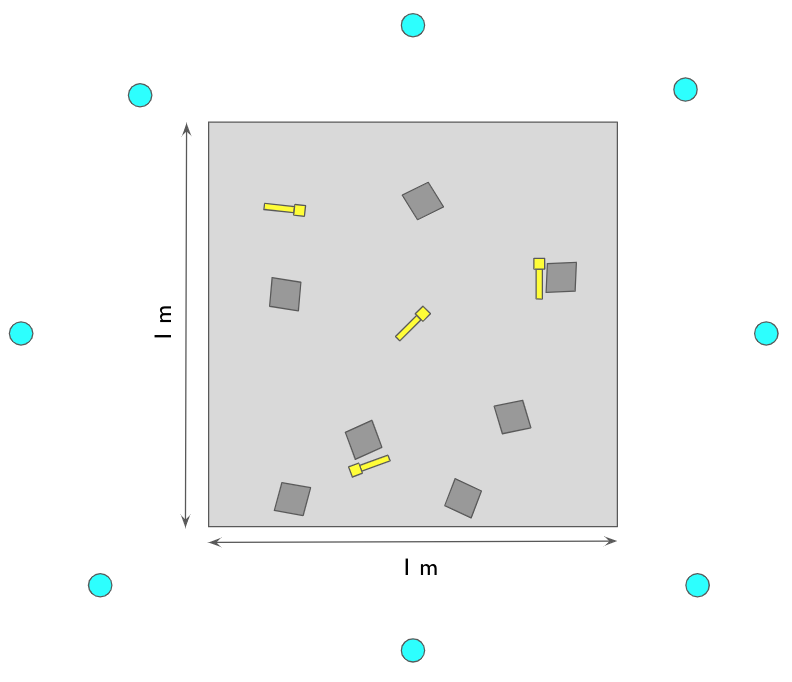}
        \label{fig:outdoordataset:circle}
    }
    \\
    \subfloat[Sample images on flagstone (left) and CFA 6 rocks (right).]{
        \includegraphics[width=0.99\columnwidth]{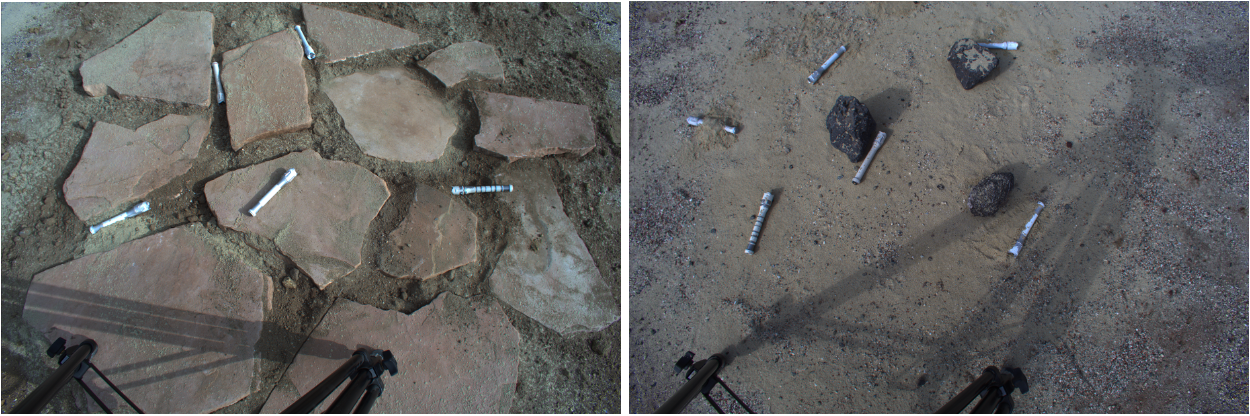}
        \label{fig:outdoordataset:fs_c6}
    }
    \\
    \subfloat[Sample images on ditch (left) and riverbed (right).]{
        \includegraphics[width=0.99\columnwidth]{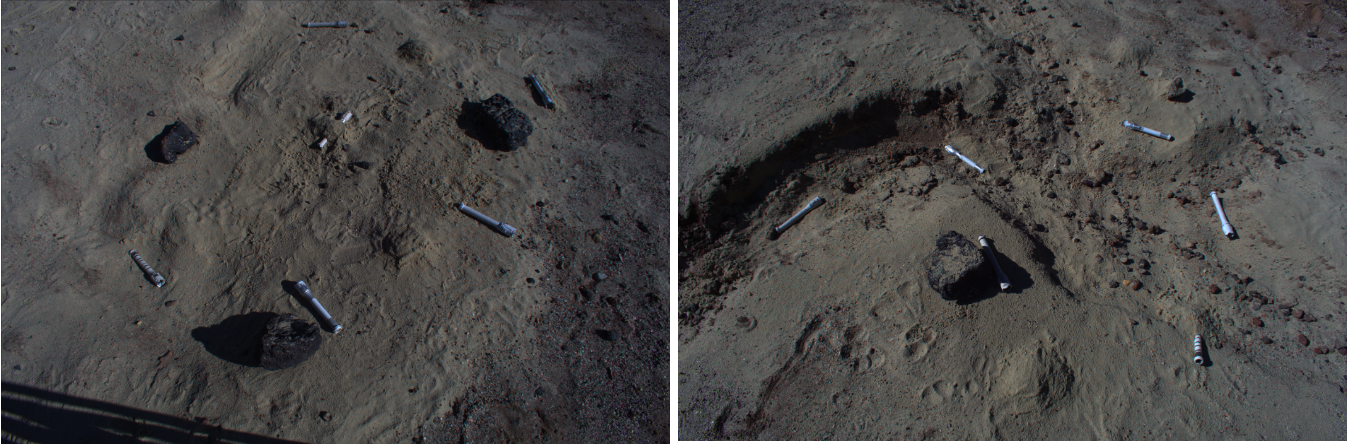}
        \label{fig:outdoordataset:ditch_riverbed}
    }
    \\
    \subfloat[Sample images with tag-mounted tube: flagstone (left) and CFA 2 (right).]{
        \includegraphics[width=0.99\columnwidth]{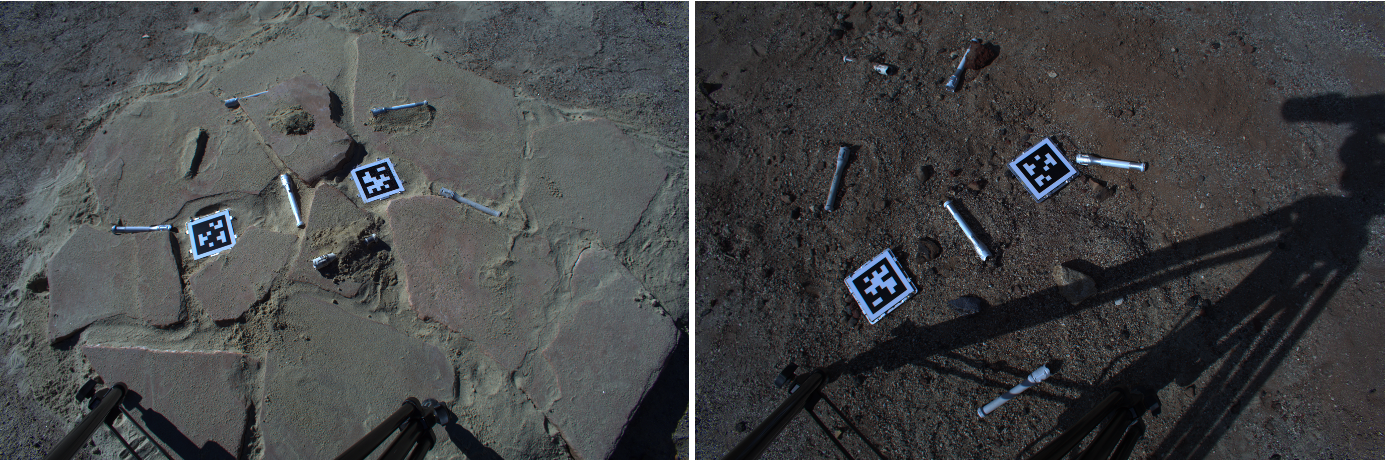}
        \label{fig:outdoordataset:fs_c2_tags}
    }
    \caption{Outdoor dataset: (a) testbed, (b) capture grid, (c-d) sample images.}
    \label{fig:outdoordataset}
\end{figure}

\subsection{Camera Acquisition Setup}
We constructed a camera acquisition setup made up of four FLIR BlackFly S cameras ($5472\times3648$, color, 77$^{\circ}$  field of view) that form two stereo pairs with baselines of $20$cm and $40$cm. The cameras, optics and the overall acquisition setup (baselines and heights) are representative of the Perseverance rover’s onboard cameras - EECAMs \cite{maki2016enhanced}. During image acquisition, we set the camera tripod at two different heights of $1$m and $2$m to simulate images acquired by HazCams and NavCams, respectively. Also, the cameras were covered by an aluminum plate serving as heat shield for extended use under sunlight (as shown in Fig \ref{fig:outdoordataset:camera_setup}). 

\subsection{Data Collection}
To evaluate the performance and robustness of our methods, we captured a dataset of outdoor images in JPL's Mars Yard. The curated set of images represents both nominal and adverse environmental conditions we expect SFR to face on Mars. An outdoor dataset provides us images of realistic scenes that contain: 1) diverse and varying terrain, informed by discussions with Mars geologists who plan where sample tubes could be dropped; 2) natural shadows and lighting that create appearance variation for the tube we are detecting. Additionally, we further vary the tube appearance by considering another adverse condition, object occlusion. Specifically, we achieve this by positioning tubes next to rocks or terrain features to induce varying levels of occlusion, and also covering tubes with dust. We vary the amount of dust coverage in a few ways: 1) sprinkling a light layer of dust; 2) building up a ``dune" on the side of a tube; 3) totally covering a partial section of the tube with a mound of sand.

\subsubsection{Ground Truth Annotations}
Finally, we enable quantitative evaluation of detection results against the dataset by providing ground truth segmentation masks with associated bounding boxes for 2D object detection benchmarking and 6DOF poses for a subset of the data, for pose estimation evaluation. The segmentation masks are manually annotated using the coco-annotator tool \cite{cocoannotator}. Ground truth 6D poses are obtained by rigidly mounting a pair of AprilTags \cite{olson_2011} to one of the tubes. Two tags (see Fig-\ref{fig:outdoordataset:fs_c2_tags}) were chosen for robustness since the occurrence of shadows sometimes hindered the tag detection algorithm. The tags are also mounted a sufficient distance away from the tube to not interfere with detection. During construction of the scene, care was also taken to cover the mount with sand so the tube appears as an isolated object.

The dataset comprises the following:
\begin{itemize}
    \item One subset of images that does not contain the ground-truth tube with the mounted AprilTags (\emph{no-tags} dataset). This set of images contains four terrain types: two terrain types constructed with consultation from the Mars geologists, and two extra ones we created based on other interesting features we saw in the Mars Yard to add more scene diversity  (see Figs.~\ref{fig:outdoordataset:fs_c6} and~\ref{fig:outdoordataset:ditch_riverbed}):

    \begin{itemize}
        \itemsep0.3em
        \item ``Flagstone'': broken stone slabs covered with a thin layer of dust emulating fractured bedrock on Mars.
        \item ``CFA6'': a rock distribution of \emph{cumulative fractional area} (CFA -- a measure of rock density) equal to 6\%. These are the smallest rocks that are still visible from orbit to guide the choice of depot location. Rocks encountered in practice would only be this big or smaller.
        \item ``Ditch'', and ``Riverbed'': These 2 terrain types were simply named for the varying levels of surface depression we observed in the terrain.
    \end{itemize}

    \item Another of subset of images that contains the tube mounted with AprilTags (\emph{with-tags} dataset). This set of images contains the three terrain types we expect to experience on Mars:
   
    \begin{itemize}
        \itemsep0.3em
        \item ``Flagstone'' and ``CFA6'': same as in the other set.
        \item ``CFA2'': small pebbles that are only visible from rover surface imagery and not orbit.
    \end{itemize} 
    
    \item Capture conditions common to both sets of images:

    \begin{itemize}
        \itemsep0.3em
        \item images taken in one of two different capture times: ``am'' ($10$am to $12$pm), or ``pm'' ($3$pm to $5$pm)
        \item 5 or 6 sample tubes with variable visibility: unoccluded, or partially occluded by rocks and/or sand and dust.
        \item images taken in one of 8 camera tripod positions on a circle around the scene, see Fig.~\ref{fig:outdoordataset:circle})
        \item 2 camera heights at each stop (1m and 2m)
        \item 4 cameras: 2 stereo pairs of baseline 20 and 40 cm
    \end{itemize}
\end{itemize}

The dataset in total contains 824 images (256 with the tag-mounted tube and 568 without), and 4852 annotated instances of tubes, out of which 256 of those have an associated 6D pose. These images were collected over 2 separate days.

%% file: sec_experiment.tex
In this section, we present experiments for both quantitative and qualitative performance analysis of the two object localizers for sample-tube localization on a Mars-like environment. All the experiments were conducted using the benchmark dataset described in Section \ref{sec:dataset}.

\subsection{Evaluation Metrics}
In order to evaluate and compare the template matching and data-driven segmentation methods presented here, we make use of some of the de facto standard statistical metrics popularized in the literature and in computer vision contests, namely average precision (AP) and average recall (AR) values and precision-recall (PR) curves.

\begin{figure}[t]
    \centering
    \includegraphics[width=0.95\linewidth]{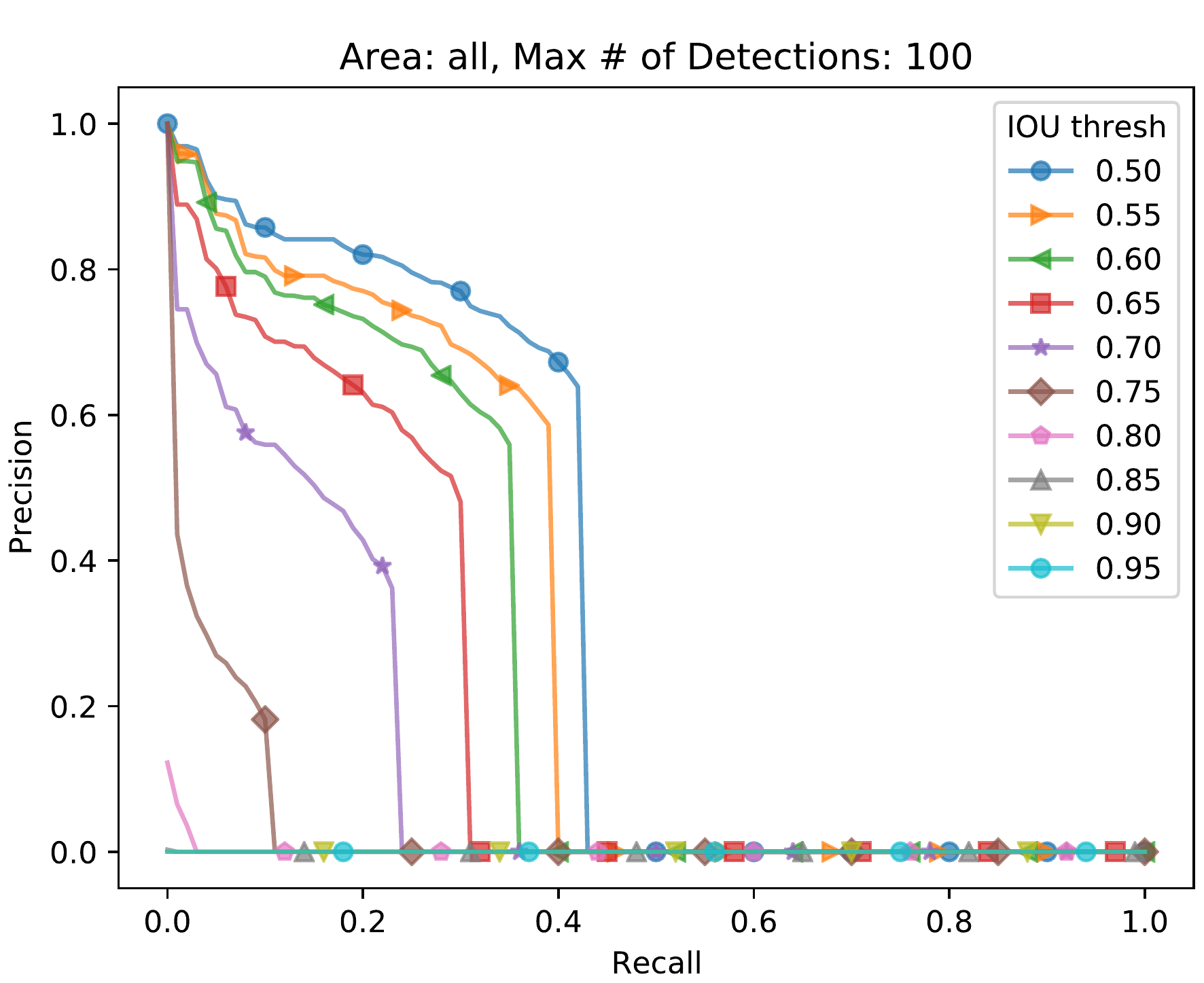}\\
    \includegraphics[width=0.95\linewidth]{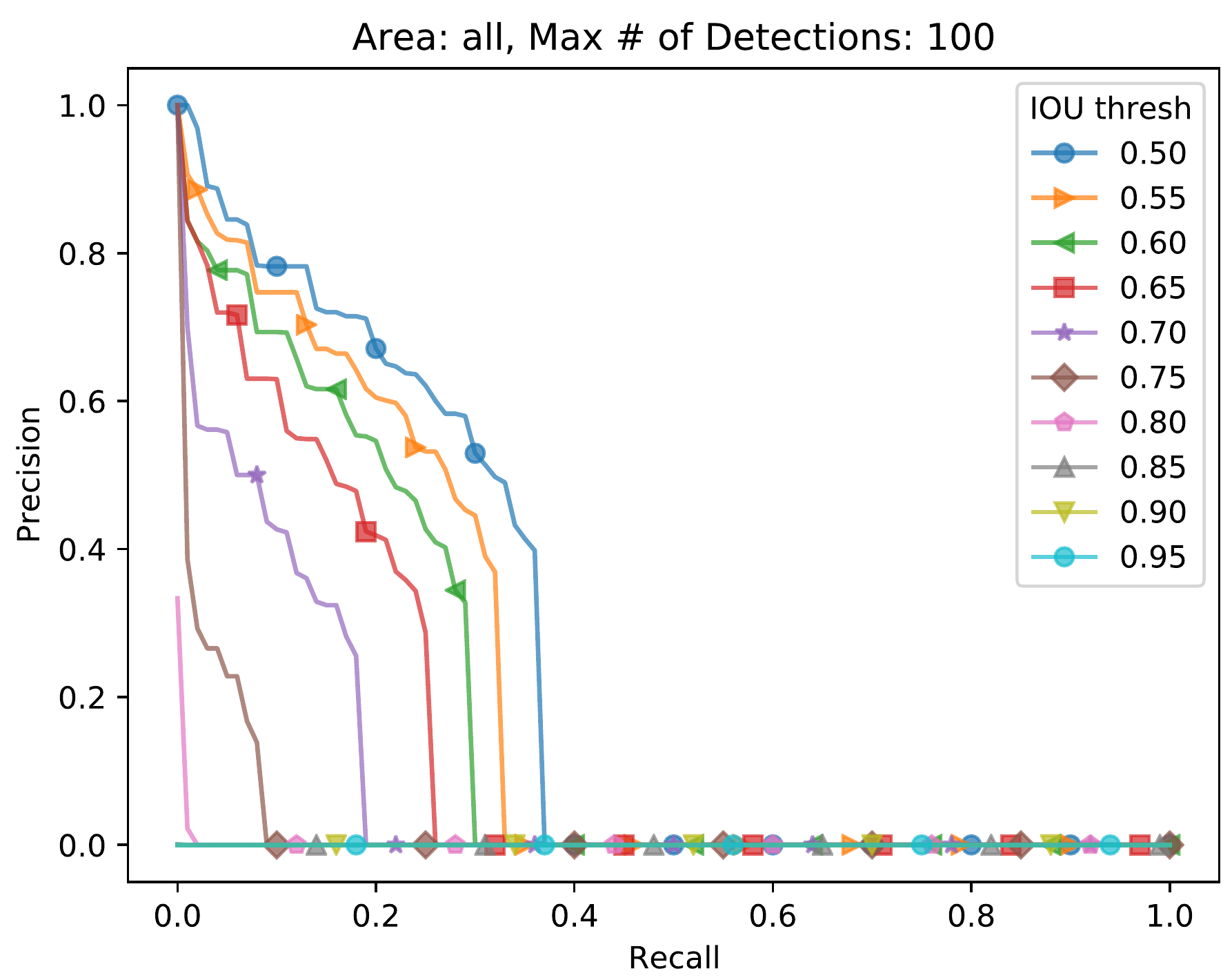}
    \caption{Precision Recall curves for the (a) no-tags and (b) with-tags datasets using the Line2D object detector.}[5pt]
    \label{fig:linemod-pr-curve}
\end{figure}

\setlength{\tabcolsep}{10pt} %
\renewcommand{\arraystretch}{1.5} %
\begin{table}[t]
    \centering
    \begin{tabular}{| c | c | c | c |}
    \hline
    \textbf{Dataset} & \textbf{Method} & \textbf{AP [.5]} & \textbf{AR [.5:.05:.95]}\\
    \hline
    no-tags & Line2D & 0.345 & 0.184\\
    \hline 
    with-tags & Line2D & 0.255 & 0.153\\
    \hline
    \end{tabular}\\[5pt]
    \caption{Quantitative results using the Line2D object detector. AP and AR IoU thresholds are shown in square brackets.}
    \label{tab:line2d}
\end{table}

\begin{figure*}[!t]
    \begin{tabular}{ccccc}
    \centering
    \includegraphics[width=0.165\linewidth]{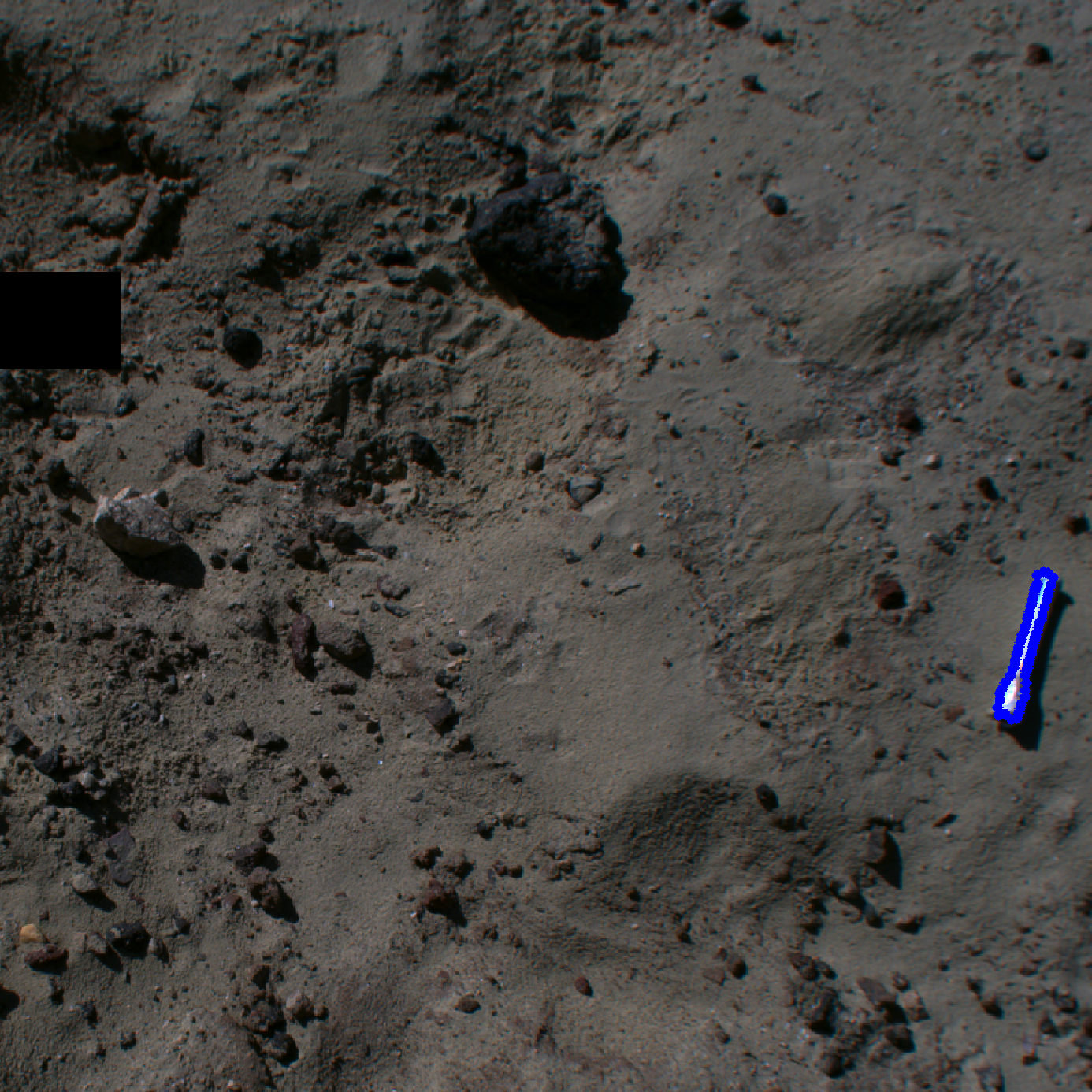}&  
    \includegraphics[width=0.165\linewidth]{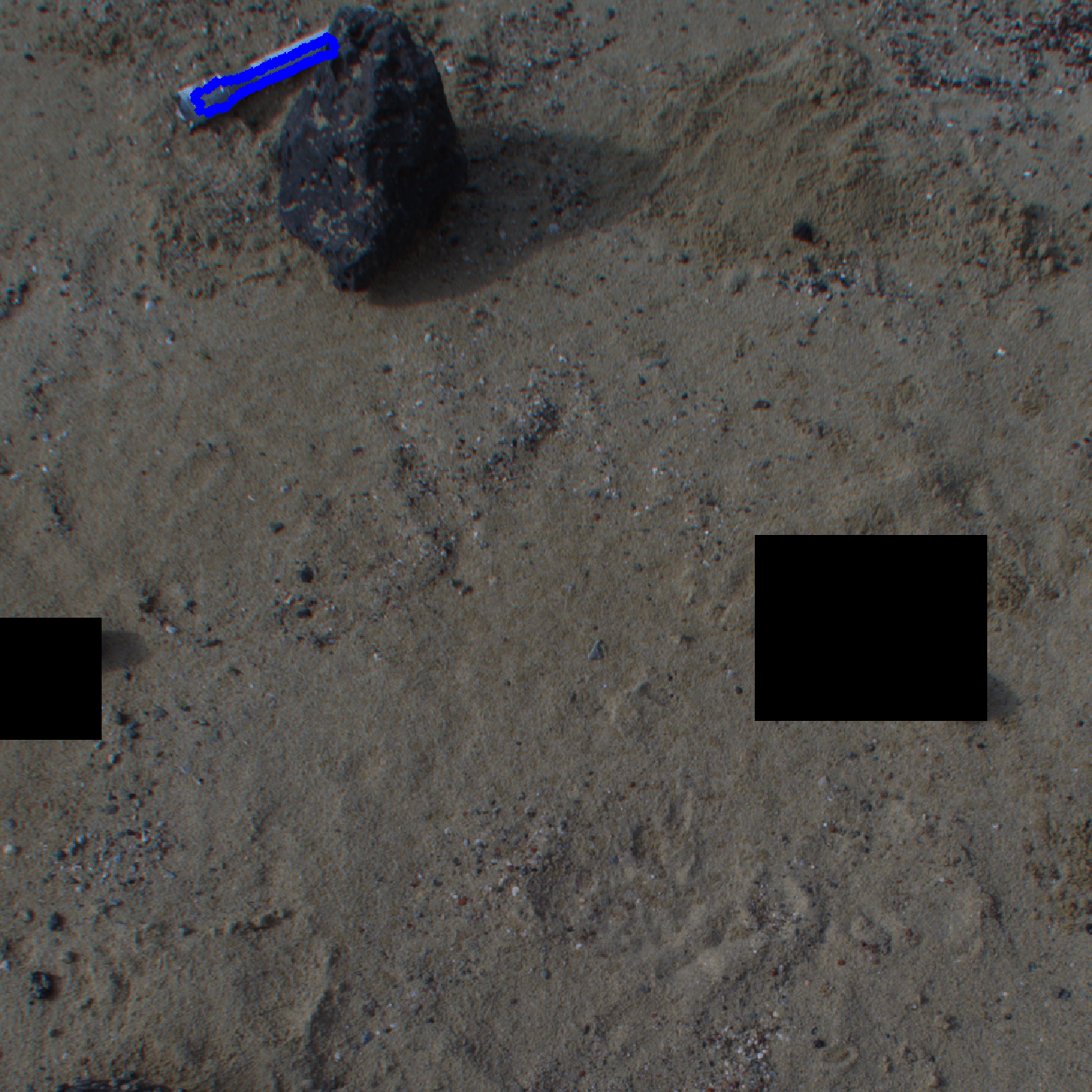}&
    \includegraphics[width=0.165\linewidth]{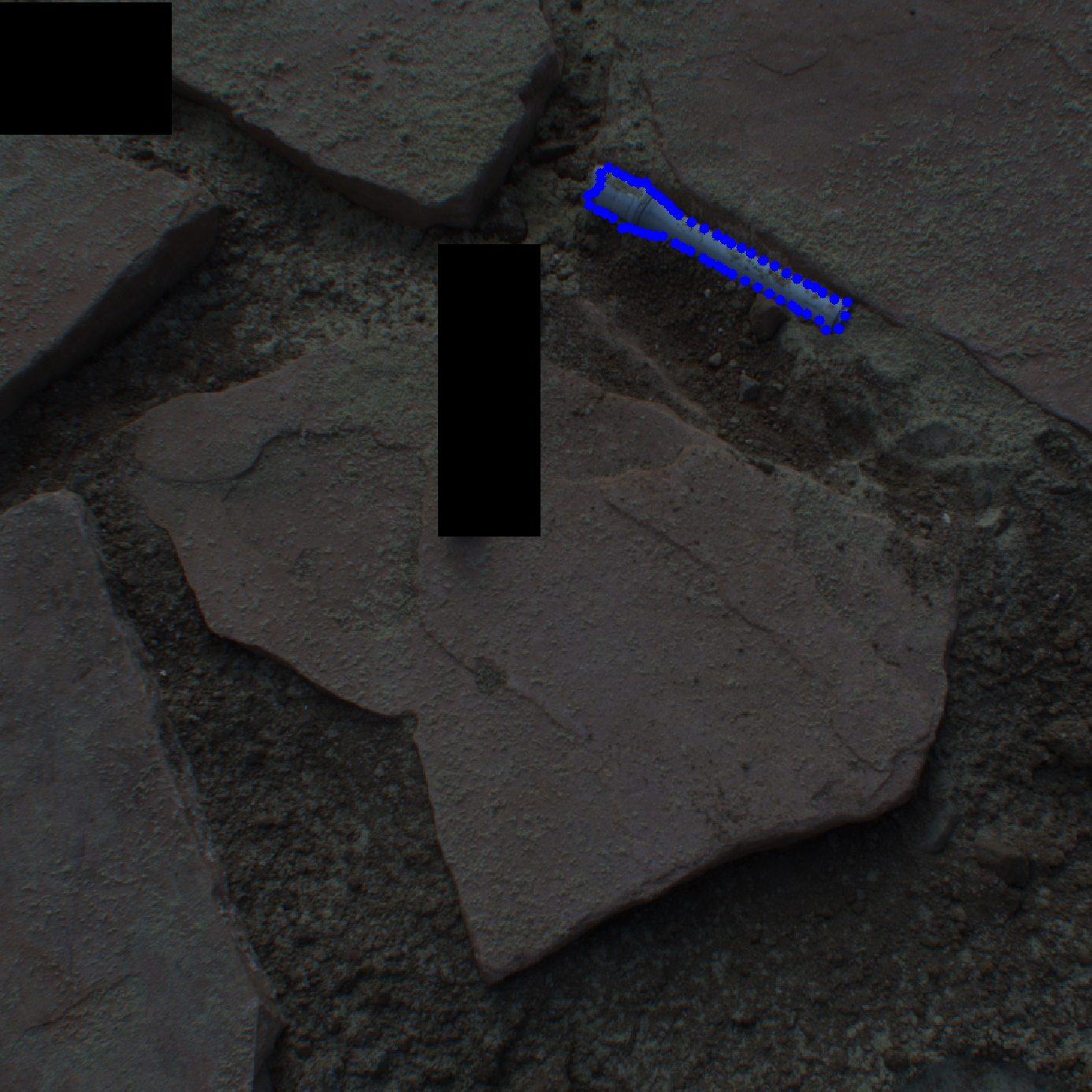}&
    \includegraphics[width=0.165\linewidth]{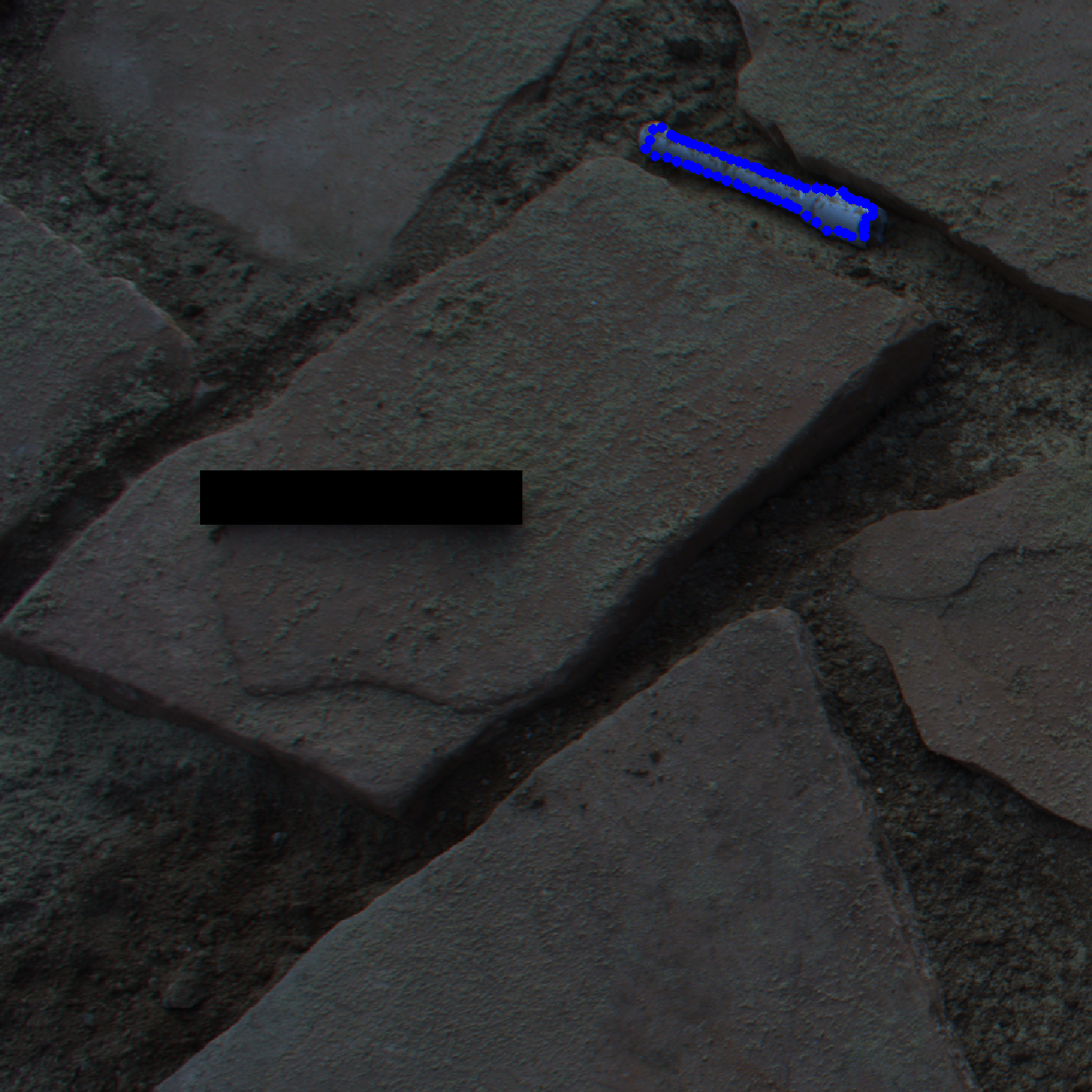}&
    \includegraphics[width=0.165\linewidth]{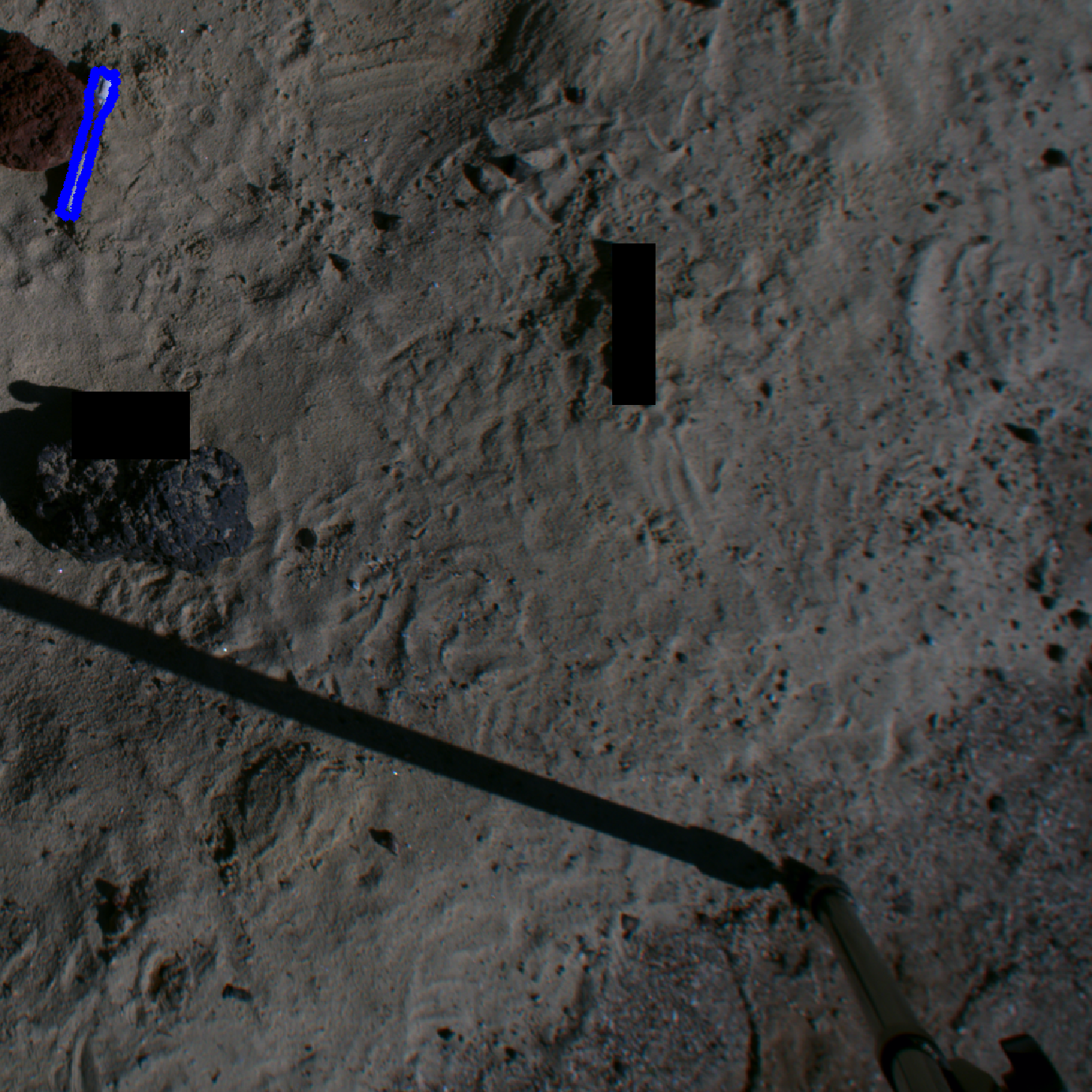}\\
    \includegraphics[width=0.165\linewidth]{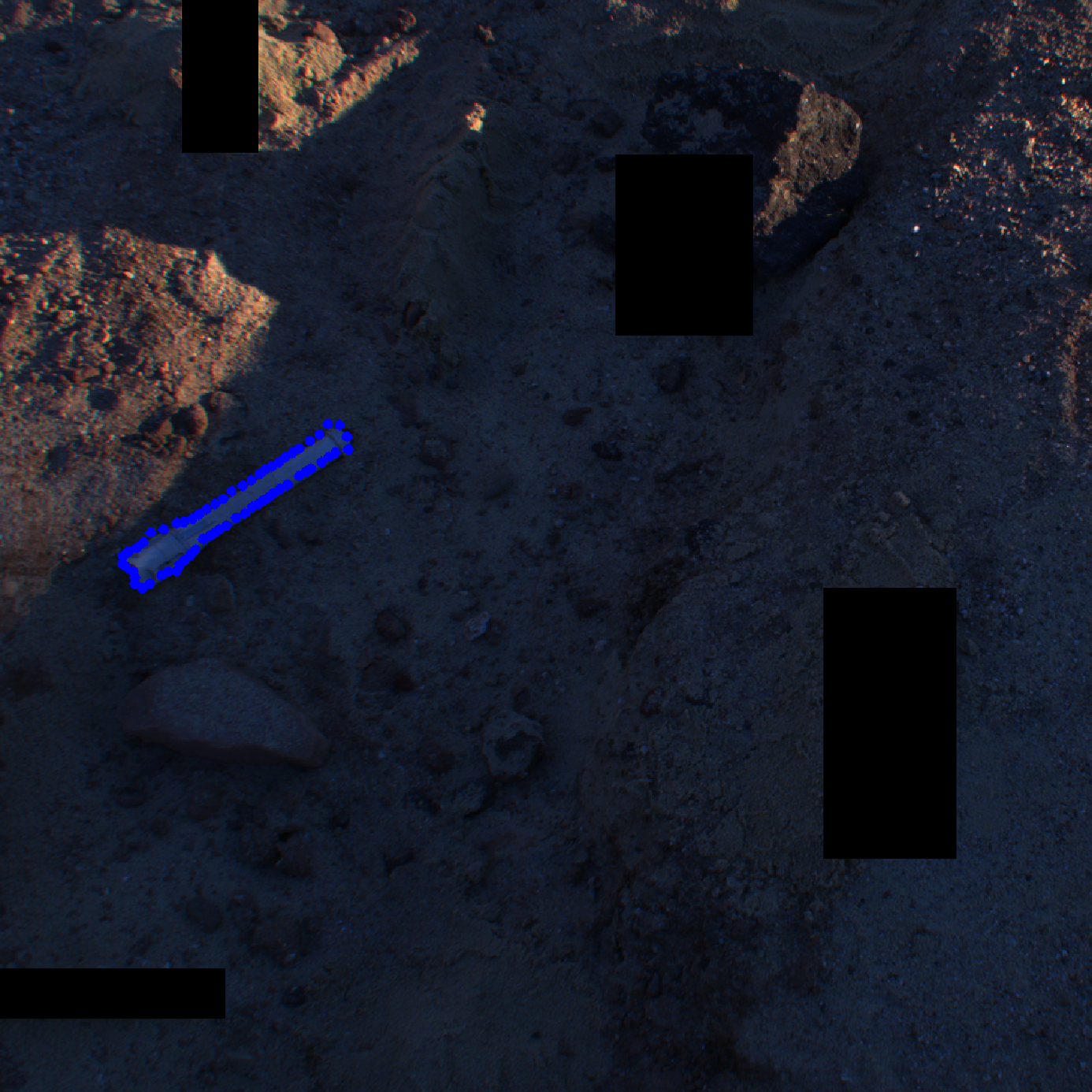} &  
    \includegraphics[width=0.165\linewidth]{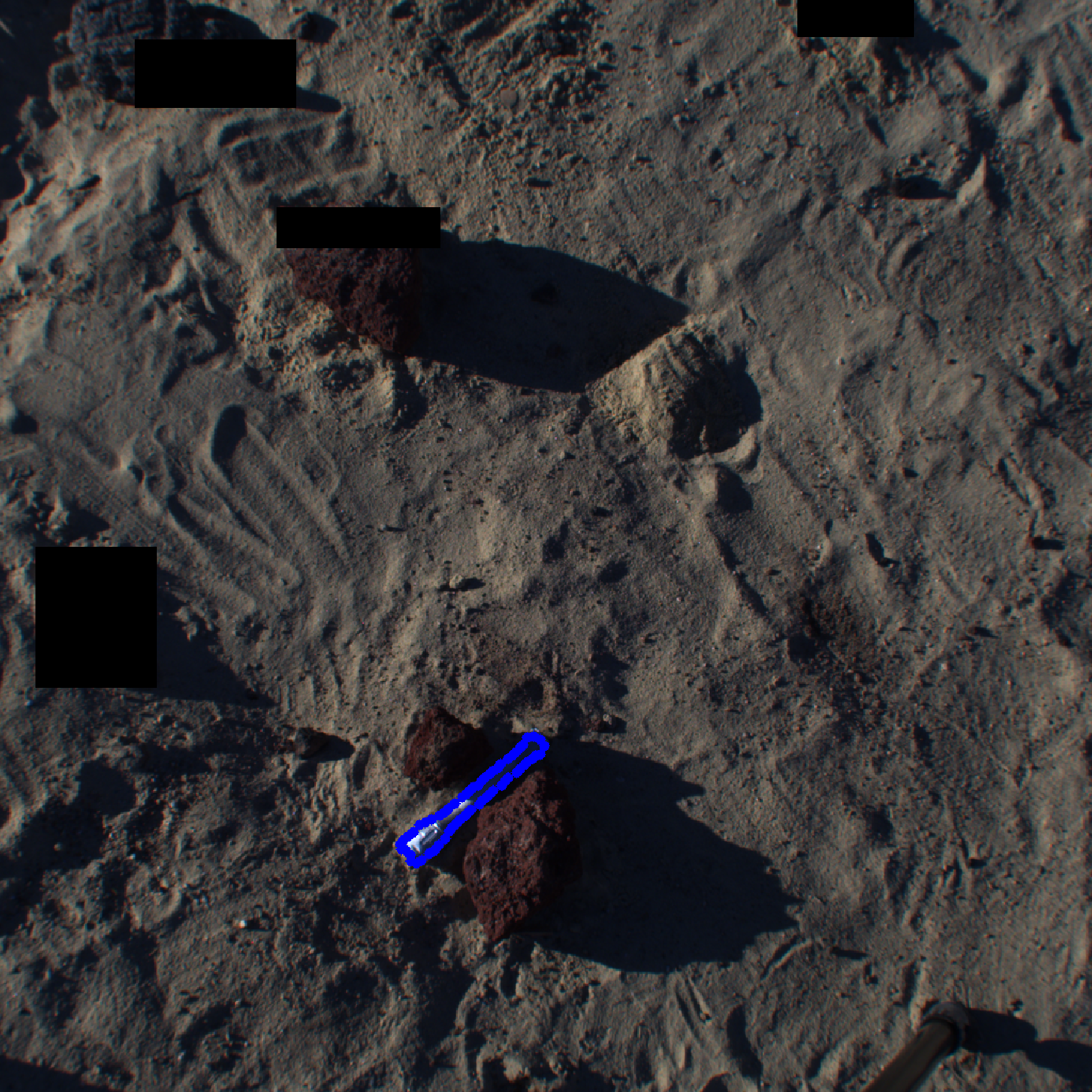}&
    \includegraphics[width=0.165\linewidth]{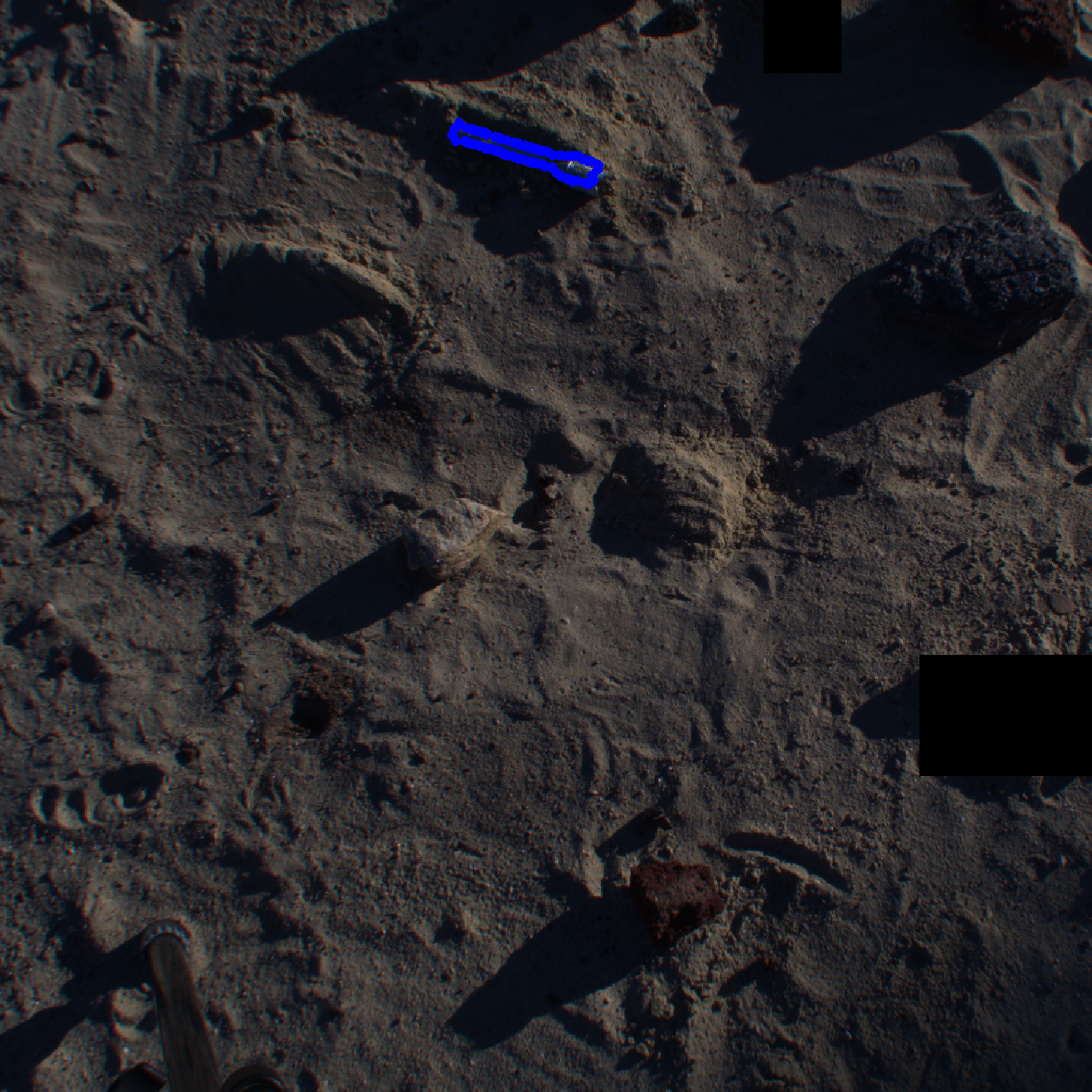}&
    \includegraphics[width=0.165\linewidth]{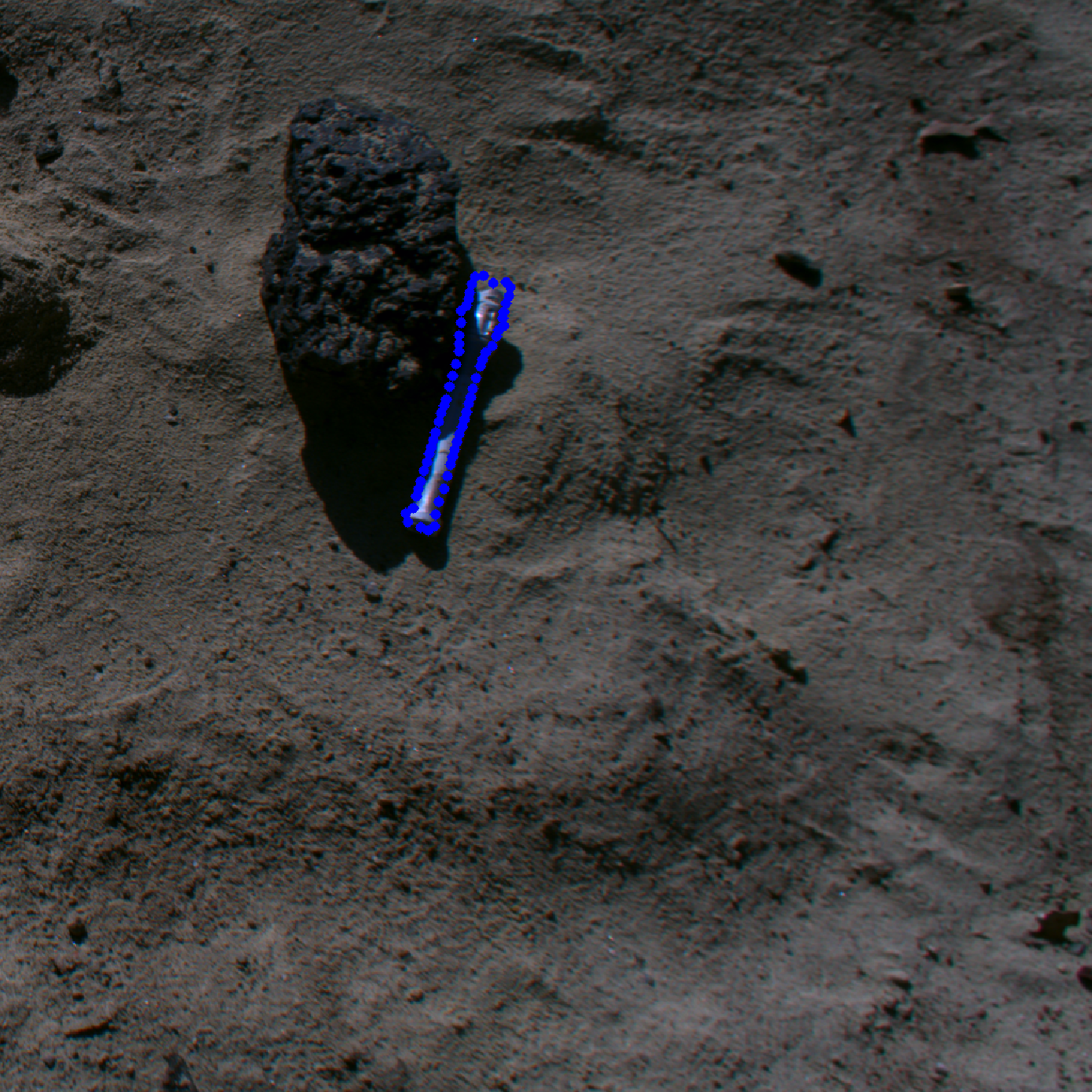}&
    \includegraphics[width=0.165\linewidth]{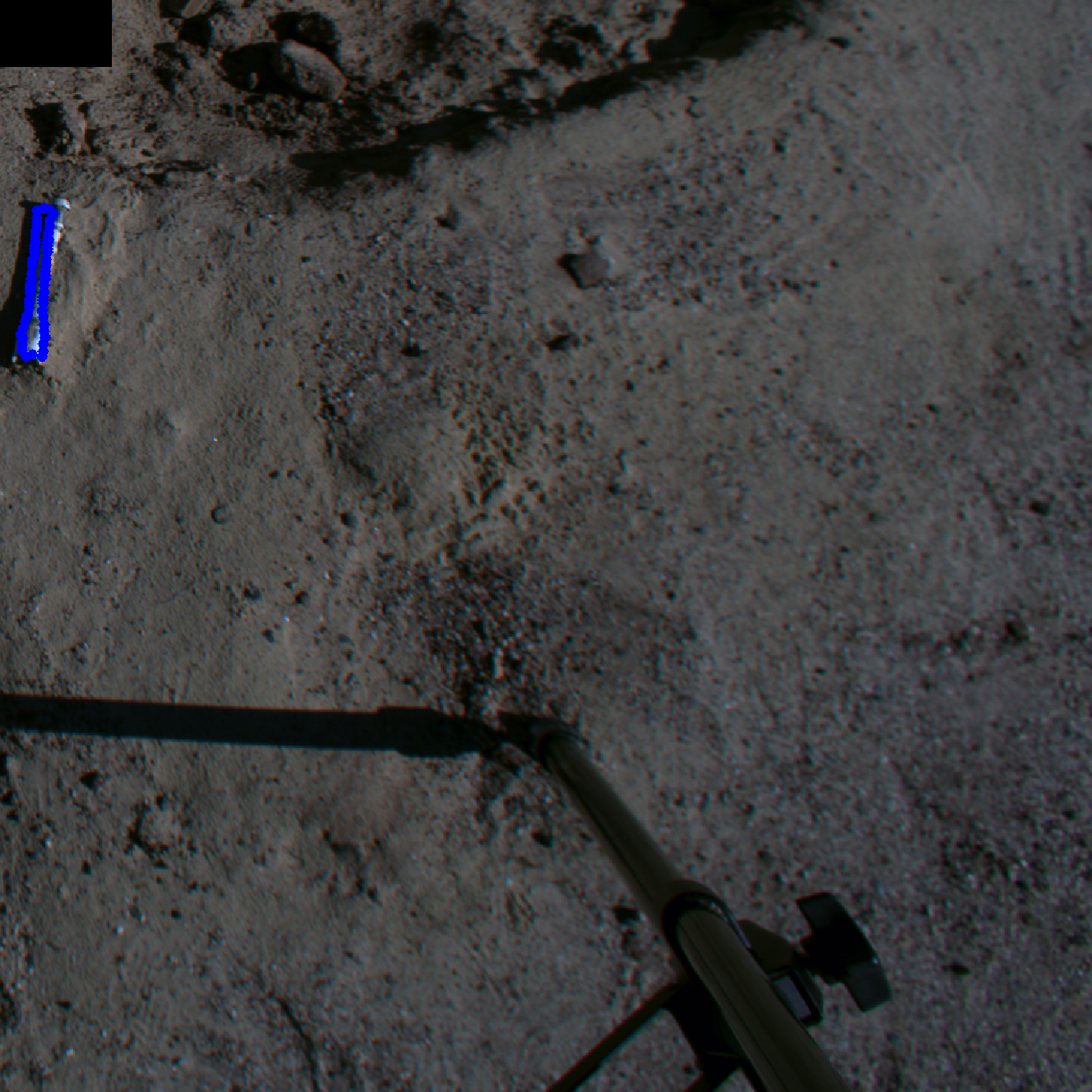}\\
    \end{tabular}
    \caption{Example true positive results for sample-tube detection using the Line2D detector.}
    \label{fig:line2d-example}
\end{figure*}

\begin{figure*}
    \begin{tabular}{ccccc}
    \centering
    \includegraphics[width=0.165\linewidth]{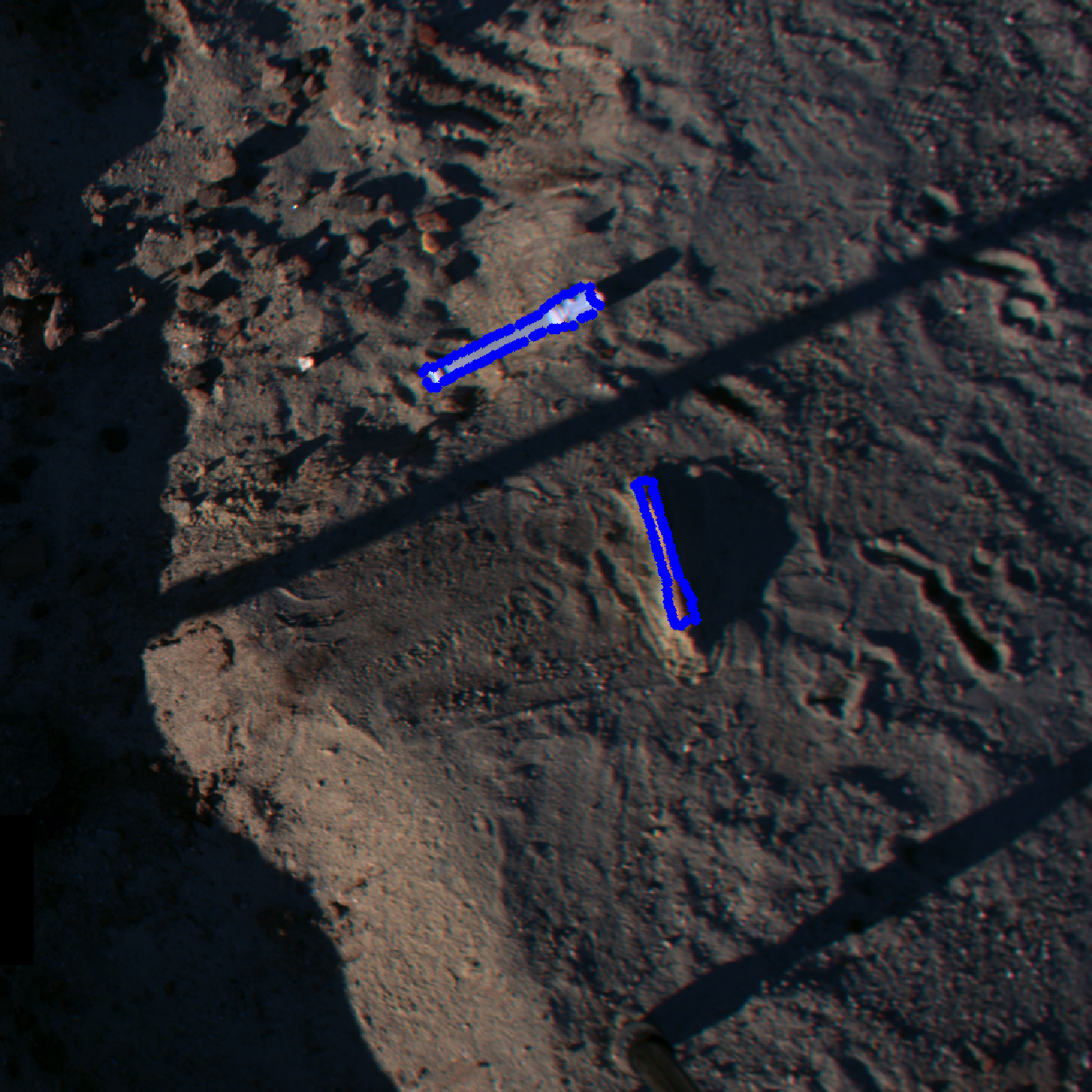}&  
    \includegraphics[width=0.165\linewidth]{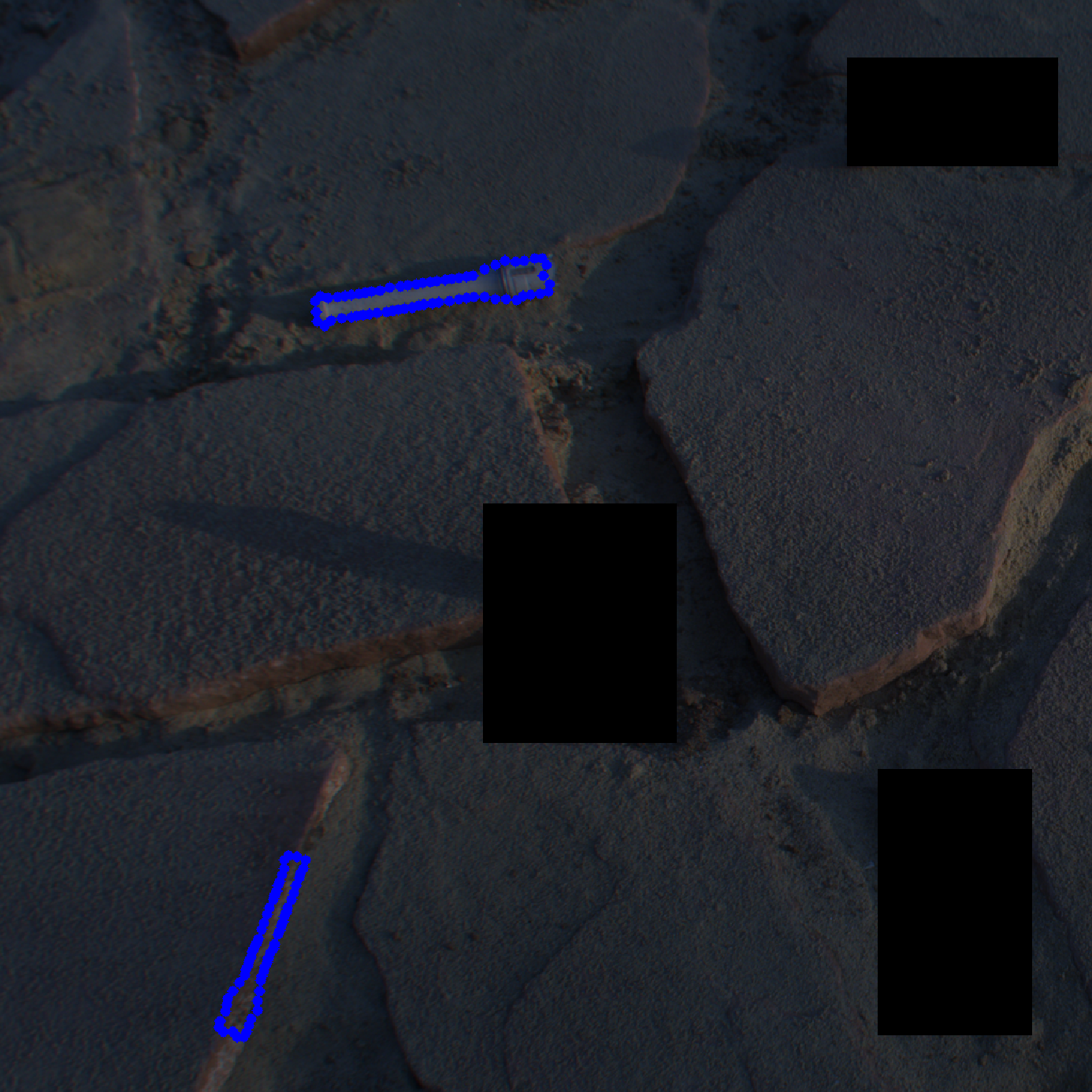}&
    \includegraphics[width=0.165\linewidth]{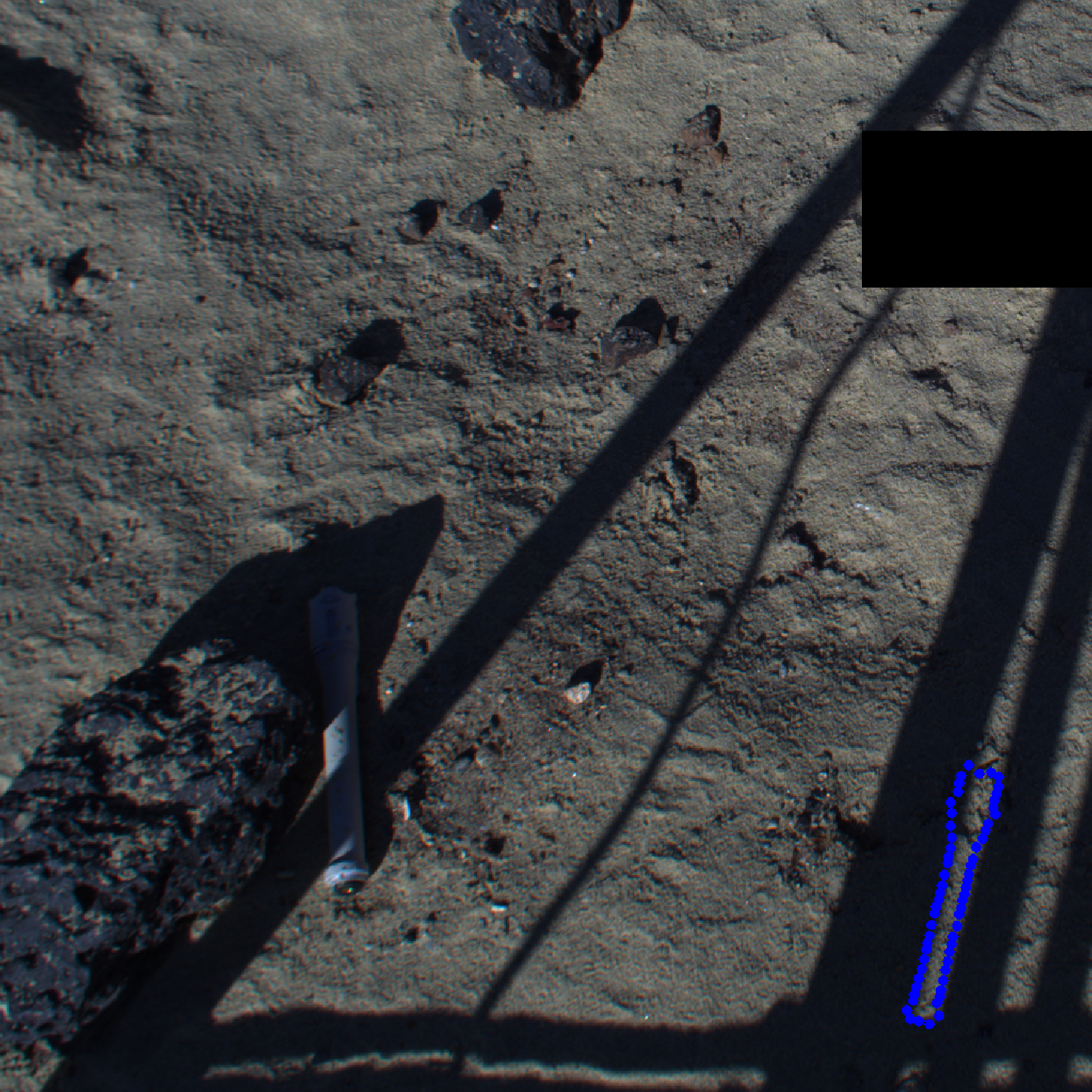}&
    \includegraphics[width=0.165\linewidth]{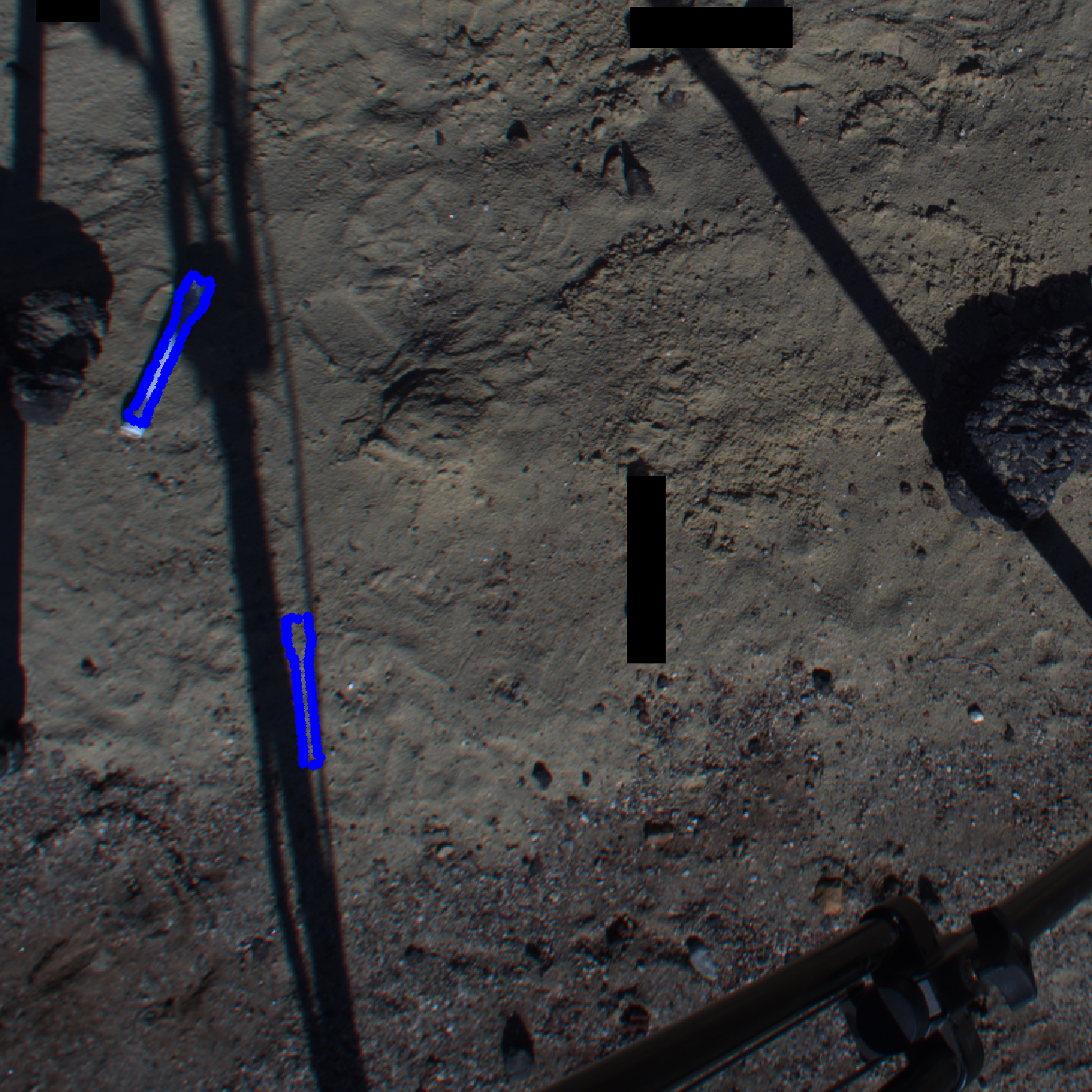}&
    \includegraphics[width=0.165\linewidth]{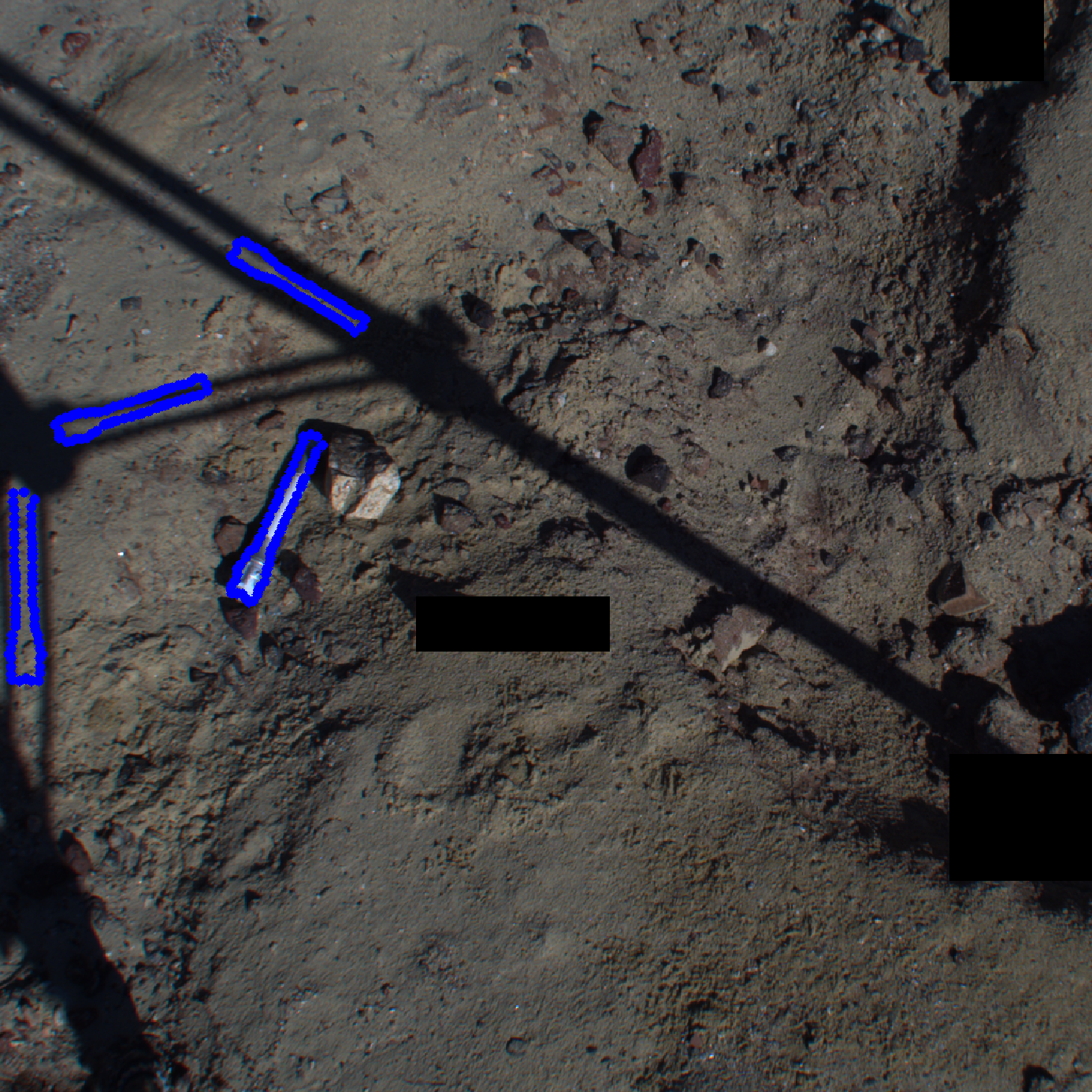}\\
    \end{tabular}
    \caption{Examples of failure cases for sample-tube detection using the the Line2D detector.}
    \label{fig:line2d-fail}
\end{figure*}

\subsubsection{Average Precision}
Various interpretations of the AP metric have been proposed over the years, particularly in the PASCAL VOC \cite{everingham2010pascal} and MS COCO \cite{lin2014microsoft} challenges. In all cases, given an intersection-over-union (IoU) threshold for determining whether a prediction should be scored as a true positive (TP) or a false positive (FP) based on the amount of overlap between the predicted mask and the ground truth mask, cumulative precision and recall values are collated in a confidence-score-ranked PR table. These values may then either be used to produce PR curves or to calculate the AP. The variation between the PASCAL VOC and MS COCO metrics lies in how the AP is calculated using these values.

The PASCAL VOC 2008 AP metric divided the recall dimension into 11 points, interpolated the precision values and averaged them, whereas PASCAL VOC 2010-2012 estimated the area under the curve. The MS COCO metric by comparison, which we use for our evaluations here, takes a 101-point AP interpolation while also evaluating the AP across multiple IoU threshold values and/or averaging over a range of thresholds. Unless otherwise stated, in the analyses below, we use the MS COCO AP at $\text{IoU}=0.5$ with all area sizes and maximum $100$ detections, and we use the MS COCO AR averaged over a range of $10$ IoU thresholds between $0.5$ and $0.95$ with a step size of $0.05$ with the same area and max detection criteria. With regard to the mission goals, since we plan to use a 6D pose estimator to refine the tube localization post-detection, we can afford to use a looser AP overlap criterion so long as precision is maximized: we would rather avoid falsely labeling tubes on the first pass than highly accurately segmenting those tubes that we do correctly detect.

\subsubsection{Precision-Recall Curves}
In the figures presented below, PR curves are shown where the interpolated precision values are used for the $101$ recall points used by the MS COCO AP metric. Plots are shown for PR curves for a range of $10$ IoU thresholds between $0.5$ and $0.95$ with a step size of $0.05$.

\subsection{Performance Analysis of Template Matching}
The Line2D detector was trained using ~7000 templates generated from the 3D model of the sample-tube. The test data consisted of images from the benchmark dataset, down-scaled to half-resolution and randomly cropped to $1024$x$1024$. The reasons to generate cropped images were two-fold. First, to match the input dimensions of the learning-based localizer for a fair comparison. Second, larger image size also increases the computational complexity of the algorithm. Table \ref{tab:line2d} shows the quantitative performance of the template matching based object localizer on the benchmark dataset using the AP and AR metrics. Furthermore, we also plot the PR curve in Figure \ref{fig:linemod-pr-curve} to demonstrate the inherent tradeoff between the precision and recall of our detector as a function of different threshold values. The overall quantitative performance of the template-based matching is encouraging, especially given the challenging and adversarial scenarios presented in our dataset. The low AP and AR values can be explained by the large number of false positives and false negatives, typical of most model-based algorithms.

Next, we qualitatively analyze the the performance of the detector as a function of different terrain types and environmental conditions. Some example results from successful detections using the Line2D detector are presented in Figure \ref{fig:line2d-example}. It can be observed that the template matching algorithm is able to robustly detect sample-tubes in a wide variety of terrain (flat ground vs. cracks of bedrock) and illumination conditions (well-lit vs partial-shadow vs completely in shadow). The shape-based similarity measure also allows the template matching to be robust to partial occlusions (see examples where tubes are places next to a rock). 

Finally we also look at some of the systematic failure cases where the template matching based detector's performance degrades. The most notable failure mode that degrades the performance of template matching was observed in the presence of other entities in the scene that have similar contour features compared to the sample-tubes. Two specific examples of this that can be observed in Figure \ref{fig:line2d-fail}, where object shadows and bedrocks - both have long edges with similar feature response to that of the long edges of the sample-tube. Another failure mode is related to occlusions. Template matching based algorithms perform poorly in the presence of significant occlusions - either from dust coverage or object occlusions.

\begin{figure}[t]
  \centering
  \includegraphics[width=\linewidth]{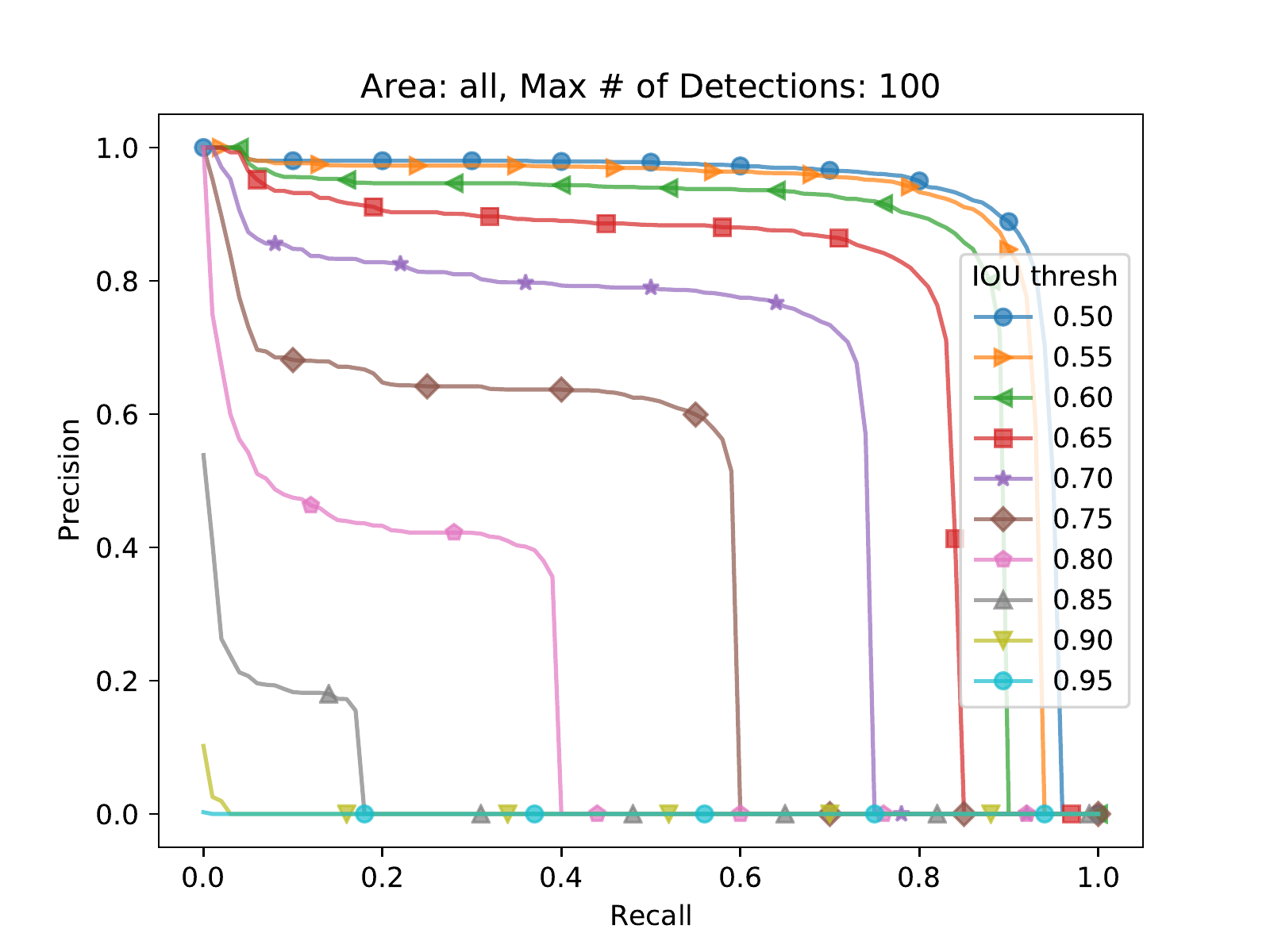}\\
  \includegraphics[width=\linewidth]{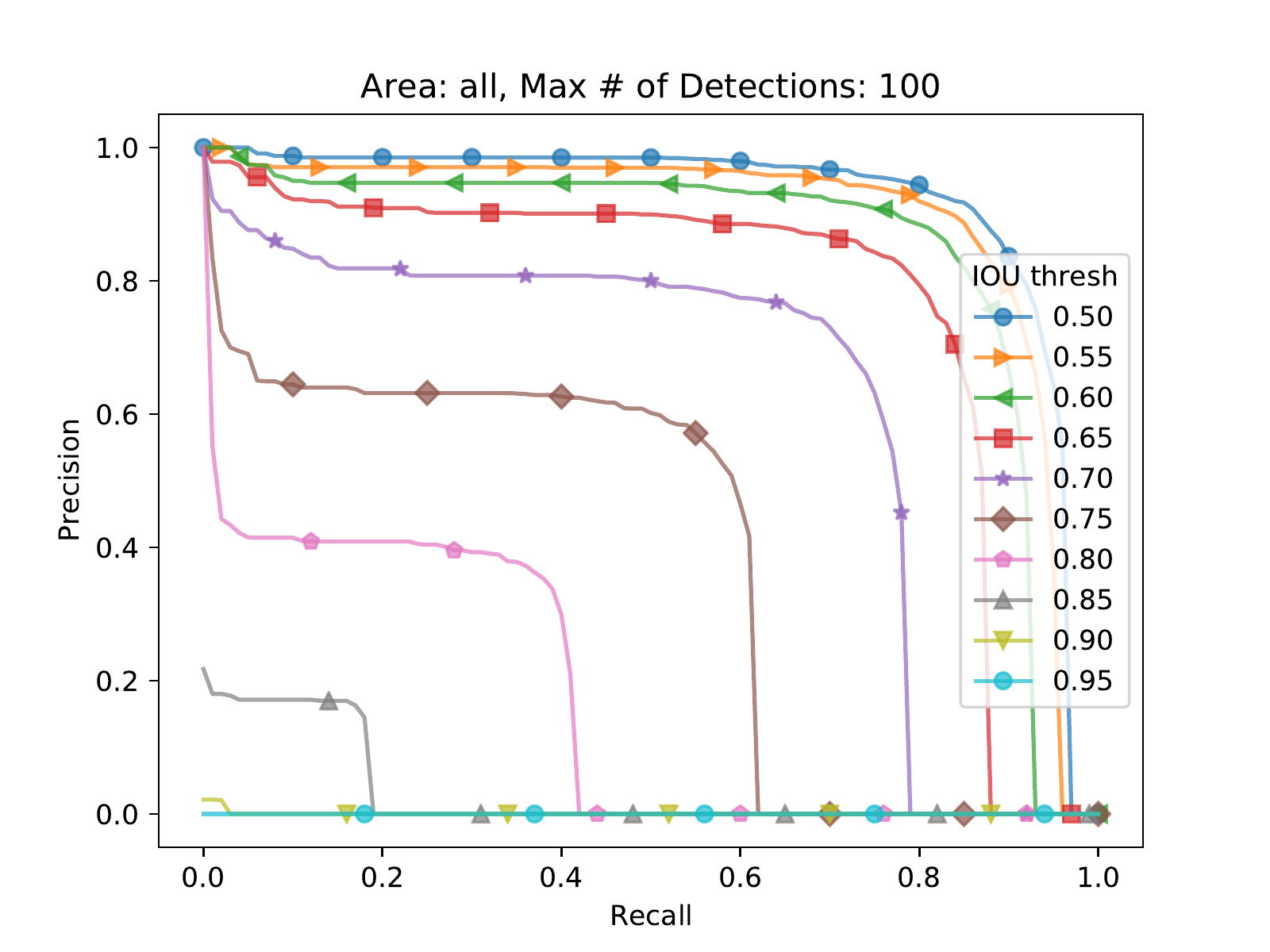}
  \caption{Precision-recall curves for the (a) no-tags and (b) with-tags datasets using the Mask R-CNN object detector.}
  \vspace{10pt}
  \label{fig:mrcnn-pr-curve}
\end{figure}

\setlength{\tabcolsep}{10pt} %
\renewcommand{\arraystretch}{1.5} %
\begin{table}[t]
  \centering
  \begin{tabular}{| c | c | c | c |}
    \hline
    \textbf{Train} & \textbf{Test} & \textbf{AP [.5]} & \textbf{AR [.5:.05:.95]}\\
    \hline
    with-tags & no-tags & 0.911 & 0.555 \\
    \hline 
    no-tags & with-tags & 0.918 & 0.575 \\
    \hline
  \end{tabular}
  \caption{Quantitative results using the Mask R-CNN object detector. AP and AR IoU thresholds are shown in square brackets.}
  \label{tab:mrcnn}
\end{table}

\subsection{Performance Analysis of Data-driven Segmentation}
Two Mask R-CNN models were trained. One was trained on the with-tags dataset and evaluated on the no-tags dataset, while the other was trained on the no-tags dataset and evaluated on the with-tags dataset. In each case the training data consists of $1200 \times 1200$ images that are converted to grayscale, box-filtered and randomly cropped to $1024 \times 1024$ to match the input size expected by the network and to allow for data augmentation. Our decision to use grayscale images reflects the intuition that color is less important for this task than edge and texture features and that the network might struggle to learn color invariance. In addition, the likelihood that the tubes would be coated with a thin layer of dust on Mars would mean that the tubes may acquire a more yellow hue than their original grey coloring that the network would be trained on. As for the box filtering, given the limited quantity of training data available, it may be helpful to prime the input to de-emphasize intensity over edge features. The test data consisted of $1024 \times 1024$ images, also converted to grayscale and box filtred, in which a whole single tube is present within the image boundaries. When the no-tags data was used for training, the with-tags data was used for testing, and vice versa.

\begin{figure*}
    \begin{tabular}{ccccc}
    \centering
    \includegraphics[width=0.165\linewidth]{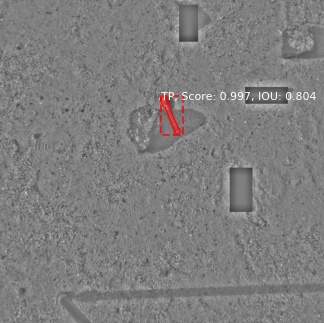}&  
    \includegraphics[width=0.165\linewidth]{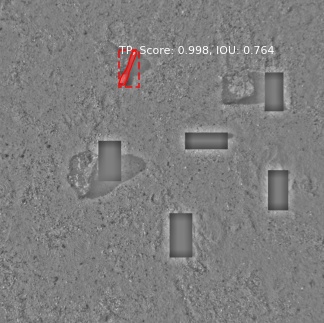}&
    \includegraphics[width=0.165\linewidth]{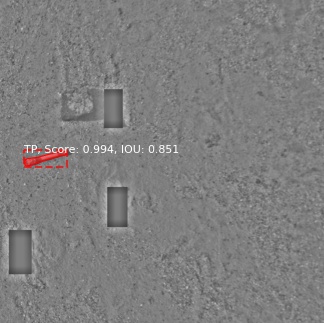}&
    \includegraphics[width=0.165\linewidth]{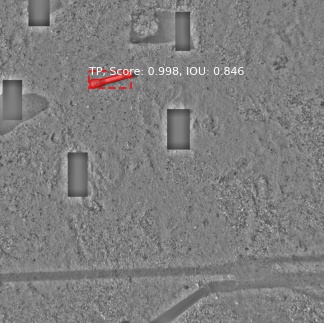}&
    \includegraphics[width=0.165\linewidth]{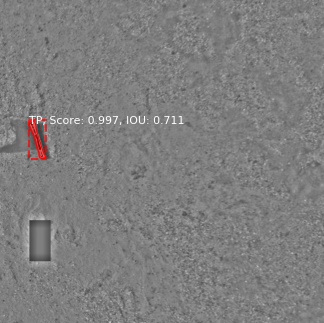}\\
    \includegraphics[width=0.165\linewidth]{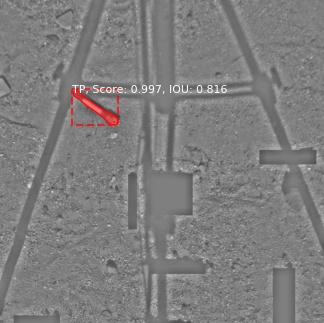}&  
    \includegraphics[width=0.165\linewidth]{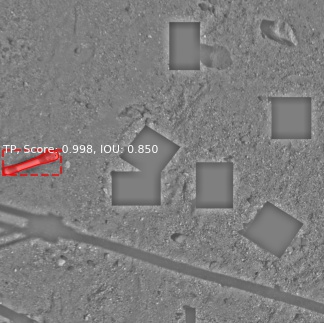}&
    \includegraphics[width=0.165\linewidth]{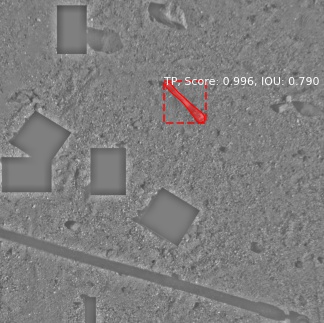}&
    \includegraphics[width=0.165\linewidth]{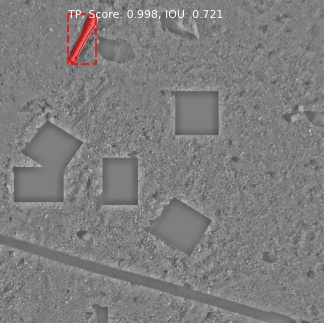}&
    \includegraphics[width=0.165\linewidth]{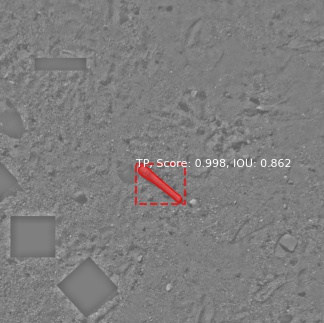}\\
    \end{tabular}
    \caption{Example true positive results for sample-tube detection using the Mask R-CNN detector. The top row shows results from the no-tags test set; the bottom row shows results from the with-tags test set. }
    \label{fig:mrcnn-example}
\end{figure*}

\begin{figure*}
    \begin{tabular}{ccccc}
    \centering
    \includegraphics[width=0.165\linewidth]{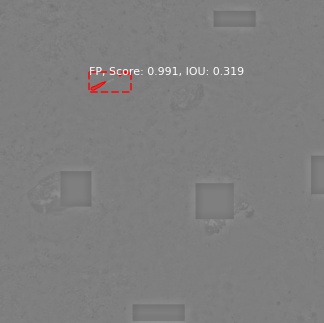}&  
    \includegraphics[width=0.165\linewidth]{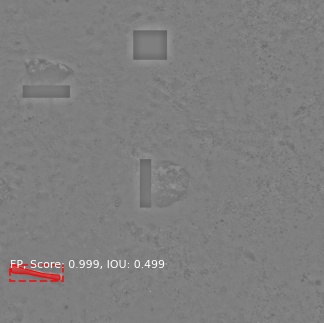}&
    \includegraphics[width=0.165\linewidth]{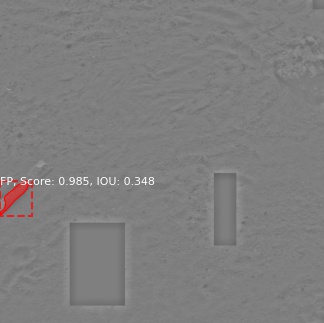}&
    \includegraphics[width=0.165\linewidth]{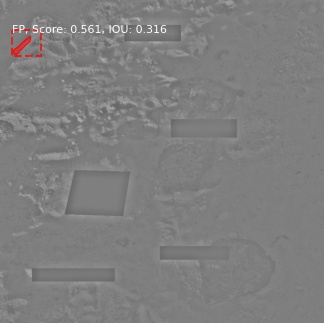}&
    \includegraphics[width=0.165\linewidth]{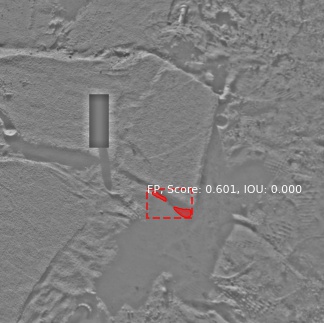}\\
    \end{tabular}
    \caption{Example false positive cases for the Mask R-CNN detector. The first three images are from the no-tags test set; the last two images are from the with-tags test set. }
    \vspace{5mm}
    \label{fig:mrcnn-fail}
\end{figure*}

Quantitative AP and AR results are presented in Table \ref{tab:mrcnn}, PR curves are presented in Figure \ref{fig:mrcnn-pr-curve}, and qualitative results samples are presented in Figures \ref{fig:mrcnn-example} and \ref{fig:mrcnn-fail}. As was expected, the data-driven learning approach of Mask R-CNN provides a substantial average increase in performance over the template-matching method both in terms of precision and recall. The PR curves also demonstrate that if we can tolerate detection overlap with ground-truth in the $0.5-0.6$ IoU threshold range, we can expect reasonably optimal precision-recall. The results samples in Figures \ref{fig:mrcnn-example} and \ref{fig:mrcnn-fail} show that Mask R-CNN can cope well even in difficult situations where object shadows overlap the tubes. Conversely, it can also produce false positives in cases where there is not enough overlap between the detection and the ground-truth or sometimes fail completely in cases where features from non-tube objects seem tube-like.

\subsection{Performance Analysis of 6D Pose Estimation}
\begin{figure*}[t]
    \centering
    \includegraphics[width=0.75\linewidth]{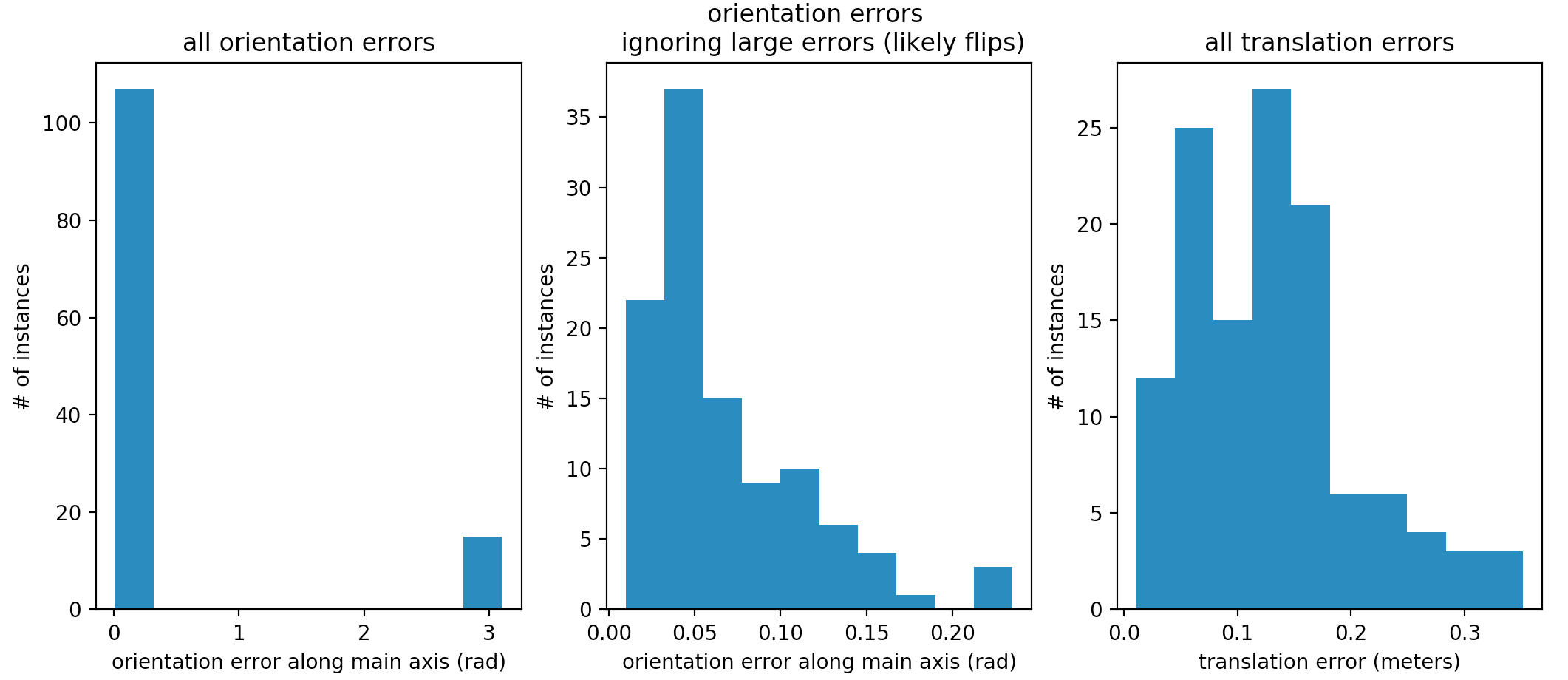}\\
    \caption{6D pose estimation results using Line2D}
    \label{fig:linemod-6d-results}
\end{figure*}

Using the \emph{with-tags} dataset with ground-truth 6D pose information, we can evaluate the pose estimation accuracy of the Line2D detector. Figure \ref{fig:linemod-6d-results} shows histograms of the orientation and translation error magnitudes. Since the detector does not know which is the tag-mounted tube, we only consider detections with $\text{IOU}>0.5$ for evaluation of the 6D pose accuracy. If there is indeed a detection in the neighborhood, there should only be one since additional ones will have been removed by non-max suppression. Moreover, we consider the orientation error in a couple of ways. First, we focus on in-plane rotation and report error along the main axis of the tube since this is the most relevant for the downstream task of tube pickup. Also, from the histograms, we see that there are a few outliers with large orientation errors. These are likely detections where the matched template is ``flipped'' 180$^{\circ}$ due to the most prominent object features being the two parallel edges that define the body of the tube. Ignoring these instances allows us to observe the nominal orientation errors. Furthermore, in practice an orientation flip should not affect ability for tube pickup.

%% file: sec_conclusion.tex
Autonomous localization and retrieval of sample-tubes for the Mars Sample Return mission is a challenging task, but one that is necessarily to accommodate the mission timeline for the Sample Fetch Rover. In this work, we studied two machine-vision based approaches to autonomously detect and localize sample-tubes. Our top-level goal was to understand the trade-off between the different classes of algorithms: model-based and data-driven. Towards this end, we also collected  a large benchmark dataset of sample-tube images, in a representative outdoor environment and annotated it with ground truth segmentation masks and locations for performance analysis. In summary, both methods have complimentary advantages. While, learning-based methods are considerably superior in terms of performance, they are fundamentally black-box from a design perspective. This presents a major challenge for Verification and Validation (V\&V) and difficulty in flight infusion. On the other hand, classical methods such as Template matching do not match the performance of their learning based counterparts, but are easier to design, implement and V\&V. In future work, we plan to validate the performance of autonomous tube-pickup through end-to-end demonstration on an analog Fetch rover.

%% file: sec_biography.tex
\begin{biographywithpic}{Shreyansh Daftry}{figs/authors/author_daftry}
    is a Robotics Technologist at NASA Jet Propulsion Laboratory, California Institute of Technology. He received his M.S. degree in Robotics from the Robotics Institute, Carnegie Mellon University in 2016, and his B.S. degree in Electronics and Communication Engineering in 2013. His research interest lies is at intersection of space technology and autonomous robotic systems, with an emphasis on machine learning applications to perception, planning and decision making. At JPL, he has worked on mission formulation for Mars Sample Return, and technology development for autonomous navigation of ground, airborne and subterranean robots.
  \end{biographywithpic}

\begin{biographywithpic}{Barry Ridge}{figs/authors/author_ridge} is a
postdoctoral scholar at the NASA Jet Propulsion Laboratory (JPL), California
Institute of Technology. He has previously held postdoctoral positions at the
Advanced Telecommunications Research Institute International, Kyoto, Japan, and
the Jo\v{z}ef Stefan Institute, Ljubljana, Slovenia. He received a B.Sc. in
Computer Applications from Dublin City University, Ireland, in 2002 and went on
to study pure mathematics at the University of St Andrews, Scotland, where he
received an M.Phil. in 2006. After being awarded a Marie Curie Fellowship to
pursue doctoral studies at the University of Ljubljana, Slovenia, with a focus
on robotic learning of object affordances, he defended his Ph.D. in 2014. His
research interests include cognitive robotics, computer vision, and machine
learning, and at JPL he is currently working on robot vision and simulation
capabilities for the Mars Sample Return and InVADER missions.
\end{biographywithpic}

\begin{biographywithpic}{William Seto}{figs/authors/author_seto}
    is a Robotics Technologist at NASA's Jet Propulsion Laboratory. He joined JPL in 2017 after receiving his M.S. in Robotic Systems Development from Carnegie Mellon's Robotics Institute. He develops software to enable autonomous capabilities in maritime and terrestrial environments. His outside interests include soccer and chicken tenders.
\end{biographywithpic}

\begin{biographywithpic}{Tu-Hoa Pham}{figs/authors/author_pham}
    is a Robotics Technologist at the NASA Jet Propulsion Laboratory, Caltech Institute of Technology, currently working on machine vision for Mars Sample Return.
    He holds a Dipl\^{o}me d'Ing\'{e}nieur in Aerospace Engineering from ISAE-SUPAERO (2013), an M.Sc. in Applied Mathematics from Universit\'{e} Paul Sabatier (2013) and a Ph.D. in Robotics from Universit\'{e} de Montpellier (2016), which he conducted at the CNRS-AIST Joint Robotics Laboratory on the topic of force sensing from vision.
    Prior to joining JPL in 2018, he spent two years as a research scientist at IBM Research Tokyo, where he worked on deep reinforcement learning for robot vision and manipulation in the real-world.
\end{biographywithpic}

\begin{biographywithpic}{Peter Ilhardt}{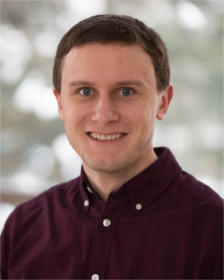}
    received his B.A. in Earth and Planetary Sciences from Northwestern University in 2013 and M.S. in Geosciences from Penn State University in 2016. He is currently a data science consultant at Capgemini and M.C.S. student at the University of Illinois. His previous work focused on spectroscopic analysis and chemical imaging of complex biogeochemical systems at Pacific Northwest National Laboratory and Penn State University. Before starting his current role, Peter interned with the Perception Systems Group at NASA JPL working on machine vision for Mars Sample Return. 
\end{biographywithpic}

\begin{biographywithpic}{Gerard Maggiolino}{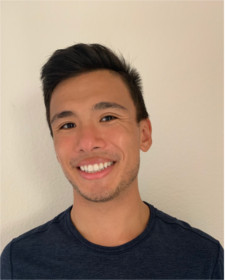}
    received a B.S. in Math-Computer Science from the University of California, San Diego in 2020 and will be attending Carnegie Mellon University for a Masters in Computer Vision in 2021. He has previously interned at NASA JPL (Pasadena), Accel Robotics (San Diego), and Elementary Robotics (Los Angeles) working on primarily Computer Vision, Machine Learning, and Software Development projects.
\end{biographywithpic}

\begin{biographywithpic}{Mark Van der Merwe}{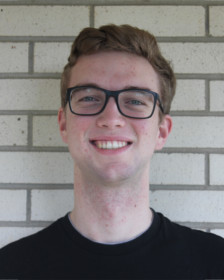}
    is a Robotics PhD student at the University of Michigan, Ann Arbor. He received his B.Sc. degree in Computer Science from the University of Utah in 2020. He interned at NASA JPL during the Summer of 2020, where he worked with the Perception Systems team on tube and rover localization for Mars Sample Return. His current research interests lie at the intersections of perception and action for robot manipulation.
\end{biographywithpic}

\begin{biographywithpic}{Alexander Brinkman}{figs/authors/author_brinkman}
    received his M.S in Robotic Systems Development from Carnegie Mellon's Robotic Institute, then joined the Robotic Manipulation and Sampling group at Jet Propulsion Laboratory in 2017.  He develops manipulation software and autonomous capabilities to enable future sampling missions to Europa, Enceladus, Mars, and comets.
\end{biographywithpic}

\begin{biographywithpic}{John Mayo}{figs/authors/author_mayo}
    is a robotics mechanical engineer in the Robotic Climbers and Grippers Group at JPL. John received a Bachelor of Science in Mechanical Engineering from Texas A\&M in 2014 and Master of Science of the same from the Massachusetts Institute of Technology in 2016.  As part of his graduate studies, John worked on hardware for the HERMES humanoid robot, developing a hybrid hand-foot device under direction of Sangbae Kim. Additionally, John co-founded and led the MIT Hyperloop Team to design and build a magnetically levitated vehicle and participated as a mentor in the new student-led shop, MIT Makerworks.
\end{biographywithpic}

\begin{biographywithpic}{Eric Kulczycki}{figs/authors/author_kulczycki}
    received a dual B.S degree in Mechanical Engineering and Aeronautical Science and Engineering from the University of California, Davis, in 2004. He received his M.S. degree in Mechanical and Aeronautical Engineering also from the University of California, Davis in 2006. He is a member of the engineering staff at the Jet Propulsion Laboratory, California Institute of Technology, where he is currently involved in Mars sample transfer chain technology development, sampling technology development for extreme environments, Mars 2020 Sample Caching Subsystem, and mechanical design of various mobility platforms. He has worked at JPL for over 15 years.
\end{biographywithpic}

\begin{biographywithpic}{Renaud Detry}{figs/authors/author_detry}  is the group leader for the Perception Systems group at NASA's Jet Propulsion Laboratory (JPL). Detry earned his Master's and Ph.D. degrees in computer engineering and robot learning from ULiege in 2006 and 2010. He served as a postdoc at KTH and ULiege between 2011 and 2015. He joined the Robotics and Mobility Section at JPL in 2016. His research interests are perception and learning for manipulation, robot grasping, and mobility, for terrestrial and planetary applications. At JPL, Detry leads the machine-vision team of the Mars Sample Return surface mission, and he conducts research in autonomous robot manipulation and mobility for Mars, Europa, Enceladus, and terrestrial applications.
\end{biographywithpic}

%% file: main.bbl
\begin{thebibliography}{10}
\providecommand{\url}[1]{#1}
\csname url@samestyle\endcsname
\providecommand{\newblock}{\relax}
\providecommand{\bibinfo}[2]{#2}
\providecommand{\BIBentrySTDinterwordspacing}{\spaceskip=0pt\relax}
\providecommand{\BIBentryALTinterwordstretchfactor}{4}
\providecommand{\BIBentryALTinterwordspacing}{\spaceskip=\fontdimen2\font plus
\BIBentryALTinterwordstretchfactor\fontdimen3\font minus
  \fontdimen4\font\relax}
\providecommand{\BIBforeignlanguage}[2]{{%
\expandafter\ifx\csname l@#1\endcsname\relax
\typeout{** WARNING: IEEEtran.bst: No hyphenation pattern has been}%
\typeout{** loaded for the language `#1'. Using the pattern for}%
\typeout{** the default language instead.}%
\else
\language=\csname l@#1\endcsname
\fi
#2}}
\providecommand{\BIBdecl}{\relax}
\BIBdecl

\bibitem{squyres2006rocks}
S.~W. Squyres, R.~E. Arvidson, D.~L. Blaney, B.~C. Clark, L.~Crumpler, W.~H.
  Farrand, S.~Gorevan, K.~E. Herkenhoff, J.~Hurowitz, A.~Kusack \emph{et~al.},
  ``Rocks of the columbia hills,'' \emph{Journal of Geophysical Research:
  Planets}, vol. 111, no.~E2, 2006.

\bibitem{grotzinger2013analysis}
J.~P. Grotzinger, ``Analysis of surface materials by the curiosity mars rover:
  Introduction,'' \emph{Science}, vol. 341, no. 6153, pp. 1475--1475, 2013.

\bibitem{board2012vision}
S.~S. Board, N.~R. Council \emph{et~al.}, \emph{Vision and voyages for
  planetary science in the decade 2013-2022}.\hskip 1em plus 0.5em minus
  0.4em\relax National Academies Press, 2012.

\bibitem{muirhead2020mars}
B.~K. Muirhead, A.~K. Nicholas, J.~Umland, O.~Sutherland, and S.~Vijendran,
  ``Mars sample return mission concept status,'' \emph{Acta Astronautica},
  2020.

\bibitem{muirhead2019sample}
B.~Muirhead, C.~Edwards, A.~Eremenko, A.~Nicholas, A.~Farrington, A.~Jackman,
  S.~Vijendran, L.~Duvet, F.~Beyer, and S.~Aziz, ``Sample retrieval lander
  concept for a potential mars sample return campaign,'' \emph{LPICo}, vol.
  2089, p. 6369, 2019.

\bibitem{mepag2008science}
M.~N. D. S.~A. Group, ``Science priorities for mars sample return,'' 2008.

\bibitem{mclennan2012planning}
S.~McLennan, M.~Sephton, D.~Beaty, M.~Hecht, B.~Pepin, I.~Leya, J.~Jones,
  B.~Weiss, M.~Race, J.~Rummel \emph{et~al.}, ``Planning for mars returned
  sample science: final report of the msr end-to-end international science
  analysis group (e2e-isag),'' \emph{Astrobiology}, vol.~12, no.~3, pp.
  175--230, 2012.

\bibitem{beaty2019a}
D.~Beaty, M.~Grady, H.~McSween, E.~Sefton-Nash, B.~Carrier, F.~Altieri,
  Y.~Amelin, E.~Ammannito, M.~Anand, L.~Benning \emph{et~al.}, ``The potential
  science and engineering value of samples delivered to earth by mars sample
  return,'' \emph{Meteoritics and Planetary Science}, vol.~54, no.~3, 2019.

\bibitem{sherwood2002mars}
B.~Sherwood, D.~B. Smith, R.~Greeley, W.~Whittaker, G.~R. Woodcock, G.~Barton,
  D.~W. Pearson, and W.~Siegfried, ``Mars sample return: Architecture and
  mission design,'' in \emph{Proceedings, IEEE Aerospace Conference},
  vol.~2.\hskip 1em plus 0.5em minus 0.4em\relax IEEE, 2002, pp. 2--536.

\bibitem{mattingly2004continuing}
R.~Mattingly, S.~Matousek, and F.~Jordan, ``Continuing evolution of mars sample
  return,'' in \emph{2004 IEEE Aerospace Conference Proceedings (IEEE Cat. No.
  04TH8720)}, vol.~1.\hskip 1em plus 0.5em minus 0.4em\relax IEEE, 2004.

\bibitem{mattingly2011a}
R.~Mattingly and L.~May, ``Mars sample return as a campaign,'' in \emph{{IEEE}
  Aerospace Conference}, 2011.

\bibitem{muirhead2020a}
B.~K. Muirhead, A.~K. Nicholas, J.~Umland, O.~Sutherland, and S.~Vijendran,
  ``Mars sample return campaign concept status,'' \emph{Acta Astronautica},
  vol. 176, pp. 131--138, 2020.

\bibitem{volpe2000technology}
R.~Volpe, E.~Baumgartner, P.~Scheaker, and S.~Hayati, ``Technology development
  and testing for enhanced mars rover sample return operations,'' in \emph{2000
  IEEE Aerospace Conference. Proceedings (Cat. No. 00TH8484)}, vol.~7.\hskip
  1em plus 0.5em minus 0.4em\relax IEEE, 2000, pp. 247--257.

\bibitem{weisbin1999autonomous}
R.~Weisbin, G.~Rodriguez, S.~Schenker, H.~Das, S.~Hayati, T.~Baumgartner,
  M.~Maimone, I.~Nesnas, and A.~Volpe, ``Autonomous rover technology for mars
  sample return,'' in \emph{Artificial Intelligence, Robotics and Automation in
  Space}, vol. 440, 1999, p.~1.

\bibitem{osinski2019canmars}
G.~R. Osinski, M.~Battler, C.~M. Caudill, R.~Francis, T.~Haltigin, V.~J.
  Hipkin, M.~Kerrigan, E.~A. Pilles, A.~Pontefract, L.~L. Tornabene
  \emph{et~al.}, ``The canmars mars sample return analogue mission,''
  \emph{Planetary and Space Science}, vol. 166, pp. 110--130, 2019.

\bibitem{younse2020concept}
P.~Younse, C.~Y. Chiu, J.~Cameron, M.~Dolci, E.~Elliot, A.~Ishigo, D.~Kogan,
  E.~Marteau, J.~Mayo, J.~Munger \emph{et~al.}, ``Concept for an on-orbit
  capture and orient module for potential mars sample return,'' in \emph{2020
  IEEE Aerospace Conference}.\hskip 1em plus 0.5em minus 0.4em\relax IEEE,
  2020, pp. 1--22.

\bibitem{perino2017evolution}
S.~Perino, D.~Cooper, D.~Rosing, L.~Giersch, Z.~Ousnamer, V.~Jamnejad,
  C.~Spurgers, M.~Redmond, M.~Lobbia, T.~Komarek \emph{et~al.}, ``The evolution
  of an orbiting sample container for potential mars sample return,'' in
  \emph{2017 IEEE Aerospace Conference}.\hskip 1em plus 0.5em minus 0.4em\relax
  IEEE, 2017, pp. 1--16.

\bibitem{edelberg2015autonomous}
K.~Edelberg, J.~Reid, R.~McCormick, L.~DuCharme, E.~Kulczycki, and P.~Backes,
  ``Autonomous localization and acquisition of a sample tube for mars sample
  return,'' in \emph{AIAA SPACE 2015 Conference and Exposition}, 2015, p. 4483.

\bibitem{papon2017a}
J.~Papon, R.~Detry, P.~Vieira, S.~Brooks, T.~Srinivasan, A.~Peterson, and
  E.~Kulczycki, ``Martian fetch: Finding and retrieving sample-tubes on the
  surface of mars,'' in \emph{{IEEE} Aerospace Conference}, 2017.

\bibitem{lee2018monocular}
B.~Lee, R.~Detry, J.~Moreno, D.~D. Lee, and E.~Kulczycki, ``Monocular visual
  pose estimation via online sampling for mars sample-tube pickup,'' in
  \emph{2018 IEEE Aerospace Conference}.\hskip 1em plus 0.5em minus 0.4em\relax
  IEEE, 2018, pp. 1--8.

\bibitem{pham2020rover}
T.-H. Pham, W.~Seto, S.~Daftry, A.~Brinkman, J.~Mayo, Y.~Cheng, C.~Padgett,
  E.~Kulczycki, and R.~Detry, ``Rover localization for tube pickup: Dataset,
  methods and validation for mars sample return planning,'' in \emph{2020 IEEE
  Aerospace Conference}.\hskip 1em plus 0.5em minus 0.4em\relax IEEE, 2020, pp.
  1--11.

\bibitem{zou2019object}
Z.~Zou, Z.~Shi, Y.~Guo, and J.~Ye, ``Object detection in 20 years: A survey,''
  \emph{arXiv preprint arXiv:1905.05055}, 2019.

\bibitem{lowe1999object}
D.~G. Lowe, ``Object recognition from local scale-invariant features,'' in
  \emph{Proceedings of the seventh IEEE international conference on computer
  vision}, vol.~2.\hskip 1em plus 0.5em minus 0.4em\relax Ieee, 1999, pp.
  1150--1157.

\bibitem{rothganger20063d}
F.~Rothganger, S.~Lazebnik, C.~Schmid, and J.~Ponce, ``3d object modeling and
  recognition using local affine-invariant image descriptors and multi-view
  spatial constraints,'' \emph{International journal of computer vision},
  vol.~66, no.~3, pp. 231--259, 2006.

\bibitem{krizhevsky2012imagenet}
A.~Krizhevsky, I.~Sutskever, and G.~E. Hinton, ``Imagenet classification with
  deep convolutional neural networks,'' in \emph{Advances in Neural Information
  Processing Systems}, 2012, pp. 1097--1105.

\bibitem{girshick2015fast}
R.~Girshick, ``Fast {{R}}-{{CNN}},'' in \emph{Proceedings of the {{IEEE
  International Conference}} on {{Computer Vision}}}, 2015, pp. 1440--1448.

\bibitem{ren2015faster}
S.~Ren, K.~He, R.~Girshick, and J.~Sun, ``Faster {{R}}-{{CNN}}: {{Towards
  Real}}-{{Time Object Detection}} with {{Region Proposal Networks}},'' in
  \emph{Advances in {{Neural Information Processing Systems}} 28}, C.~Cortes,
  N.~D. Lawrence, D.~D. Lee, M.~Sugiyama, and R.~Garnett, Eds.\hskip 1em plus
  0.5em minus 0.4em\relax {Curran Associates, Inc.}, 2015, pp. 91--99.

\bibitem{long2015fully}
J.~Long, E.~Shelhamer, and T.~Darrell, ``Fully convolutional networks for
  semantic segmentation,'' in \emph{Proceedings of the {{IEEE Conference}} on
  {{Computer Vision}} and {{Pattern Recognition}}}, 2015, pp. 3431--3440.

\bibitem{he2017mask}
K.~He, G.~Gkioxari, P.~Dollar, and R.~Girshick, ``Mask {{R}}-{{CNN}},'' in
  \emph{Proceedings of the {{IEEE International Conference}} on {{Computer
  Vision}}}, 2017, pp. 2961--2969.

\bibitem{ono2020maars}
M.~Ono, B.~Rothrock, K.~Otsu, S.~Higa, Y.~Iwashita, A.~Didier, T.~Islam,
  C.~Laporte, V.~Sun, K.~Stack \emph{et~al.}, ``Maars: Machine learning-based
  analytics for automated rover systems,'' in \emph{2020 IEEE Aerospace
  Conference}.\hskip 1em plus 0.5em minus 0.4em\relax IEEE, 2020, pp. 1--17.

\bibitem{abcouwer2020machine}
N.~Abcouwer, S.~Daftry, S.~Venkatraman, T.~del Sesto, O.~Toupet, R.~Lanka,
  J.~Song, Y.~Yue, and M.~Ono, ``Machine learning based path planning for
  improved rover navigation (pre-print version),'' in \emph{2021 IEEE Aerospace
  Conference}.\hskip 1em plus 0.5em minus 0.4em\relax IEEE, 2021.

\bibitem{costa20003d}
M.~S. Costa and L.~G. Shapiro, ``3d object recognition and pose with relational
  indexing,'' \emph{Computer Vision and Image Understanding}, vol.~79, no.~3,
  pp. 364--407, 2000.

\bibitem{david2005object}
P.~David and D.~DeMenthon, ``Object recognition in high clutter images using
  line features,'' in \emph{Tenth IEEE International Conference on Computer
  Vision (ICCV'05) Volume 1}, vol.~2.\hskip 1em plus 0.5em minus 0.4em\relax
  IEEE, 2005, pp. 1581--1588.

\bibitem{weiss2001model}
I.~Weiss and M.~Ray, ``Model-based recognition of 3d objects from single
  images,'' \emph{IEEE Transactions on Pattern Analysis and Machine
  Intelligence}, vol.~23, no.~2, pp. 116--128, 2001.

\bibitem{cyr2004similarity}
C.~M. Cyr and B.~B. Kimia, ``A similarity-based aspect-graph approach to 3d
  object recognition,'' \emph{International Journal of Computer Vision},
  vol.~57, no.~1, pp. 5--22, 2004.

\bibitem{eggert1993scale}
D.~W. Eggert, K.~W. Bowyer, C.~R. Dyer, H.~I. Christensen, and D.~B. Goldgof,
  ``The scale space aspect graph,'' \emph{IEEE Transactions on Pattern Analysis
  and Machine Intelligence}, vol.~15, no.~11, pp. 1114--1130, 1993.

\bibitem{ulrich2011combining}
M.~Ulrich, C.~Wiedemann, and C.~Steger, ``Combining scale-space and
  similarity-based aspect graphs for fast 3d object recognition,'' \emph{IEEE
  transactions on pattern analysis and machine intelligence}, vol.~34, no.~10,
  pp. 1902--1914, 2011.

\bibitem{hinterstoisser2011gradient}
S.~Hinterstoisser, C.~Cagniart, S.~Ilic, P.~Sturm, N.~Navab, P.~Fua, and
  V.~Lepetit, ``Gradient response maps for real-time detection of textureless
  objects,'' \emph{IEEE transactions on pattern analysis and machine
  intelligence}, vol.~34, no.~5, pp. 876--888, 2011.

\bibitem{hinterstoisser2012model}
S.~Hinterstoisser, V.~Lepetit, S.~Ilic, S.~Holzer, G.~Bradski, K.~Konolige, and
  N.~Navab, ``Model based training, detection and pose estimation of
  texture-less 3d objects in heavily cluttered scenes,'' in \emph{Asian
  conference on computer vision}.\hskip 1em plus 0.5em minus 0.4em\relax
  Springer, 2012, pp. 548--562.

\bibitem{cai2013fast}
H.~Cai, T.~Werner, and J.~Matas, ``Fast detection of multiple textureless 3-d
  objects,'' in \emph{International Conference on Computer Vision
  Systems}.\hskip 1em plus 0.5em minus 0.4em\relax Springer, 2013, pp.
  103--112.

\bibitem{tombari2010a}
F.~Tombari, S.~Salti, and L.~Di~Stefano, ``Unique signatures of histograms for
  local surface description,'' in \emph{European Conference on Computer
  Vision}, 2010, pp. 356--369.

\bibitem{tsai2018real}
C.-Y. Tsai, C.-C. Yu \emph{et~al.}, ``Real-time textureless object detection
  and recognition based on an edge-based hierarchical template matching
  algorithm,'' \emph{Journal of Applied Science and Engineering}, vol.~21,
  no.~2, pp. 229--240, 2018.

\bibitem{do2018deep}
T.-T. Do, M.~Cai, T.~Pham, and I.~Reid, ``Deep-6dpose: Recovering 6d object
  pose from a single rgb image,'' \emph{arXiv preprint arXiv:1802.10367}, 2018.

\bibitem{kehl2017ssd}
W.~Kehl, F.~Manhardt, F.~Tombari, S.~Ilic, and N.~Navab, ``Ssd-6d: Making
  rgb-based 3d detection and 6d pose estimation great again,'' in
  \emph{Proceedings of the IEEE International Conference on Computer Vision},
  2017, pp. 1521--1529.

\bibitem{xiang2017posecnn}
Y.~Xiang, T.~Schmidt, V.~Narayanan, and D.~Fox, ``Posecnn: A convolutional
  neural network for 6d object pose estimation in cluttered scenes,''
  \emph{arXiv preprint arXiv:1711.00199}, 2017.

\bibitem{pham2021rover}
T.-H. Pham, W.~Seto, S.~Daftry, B.~Ridge, J.~Hansen, T.~Thrush, J.~Mayo,
  Y.~Cheng, C.~Padgett, E.~Kulczycki, and R.~Detry, ``Rover {{Navigation}} for
  {{Mars Sample Return Planning}}: {{Relocalization}} in {{Changing
  Environments}} by {{Virtual Template Synthesis}} and {{Matching}},'' in
  \emph{2021 {{IEEE International Conference}} on {{Robotics}} and
  {{Automation}} ({{ICRA}})}, {Xi'an, China}, May 2021 ({{Under Review}}).

\bibitem{francis2017a}
R.~Francis, T.~Estlin, G.~Doran, S.~Johnstone, D.~Gaines, V.~Verma, M.~Burl,
  J.~Frydenvang, S.~Montaño, R.~Wiens \emph{et~al.}, ``Aegis autonomous
  targeting for chemcam on mars science laboratory: Deployment and results of
  initial science team use,'' \emph{Science Robotics}, vol.~2, no.~7, 2017.

\bibitem{kim2005a}
W.~Kim, R.~Steele, A.~Ansar, K.~Ali, and I.~Nesnas, ``Rover-based visual target
  tracking validation and mission infusion,'' in \emph{Space 2005}, 2005.

\bibitem{lowe2004distinctive}
D.~G. Lowe, ``Distinctive image features from scale-invariant keypoints,''
  \emph{International journal of computer vision}, vol.~60, no.~2, pp. 91--110,
  2004.

\bibitem{steger2002occlusion}
C.~Steger, ``Occlusion, clutter, and illumination invariant object
  recognition,'' \emph{International Archives of Photogrammetry Remote Sensing
  and Spatial Information Sciences}, vol.~34, no. 3/A, pp. 345--350, 2002.

\bibitem{rios2013discriminatively}
R.~Rios-Cabrera and T.~Tuytelaars, ``Discriminatively trained templates for 3d
  object detection: A real time scalable approach,'' in \emph{Proceedings of
  the IEEE international conference on computer vision}, 2013, pp. 2048--2055.

\bibitem{he2016deep}
K.~He, X.~Zhang, S.~Ren, and J.~Sun, ``Deep {{Residual Learning}} for {{Image
  Recognition}},'' in \emph{Proceedings of the {{IEEE Conference}} on
  {{Computer Vision}} and {{Pattern Recognition}}}, 2016, pp. 770--778.

\bibitem{xie2017aggregated}
S.~Xie, R.~Girshick, P.~Dollar, Z.~Tu, and K.~He, ``Aggregated {{Residual
  Transformations}} for {{Deep Neural Networks}},'' in \emph{Proceedings of the
  {{IEEE Conference}} on {{Computer Vision}} and {{Pattern Recognition}}},
  2017, pp. 1492--1500.

\bibitem{lin2017feature}
T.-Y. Lin, P.~Dollar, R.~Girshick, K.~He, B.~Hariharan, and S.~Belongie,
  ``Feature {{Pyramid Networks}} for {{Object Detection}},'' in
  \emph{Proceedings of the {{IEEE Conference}} on {{Computer Vision}} and
  {{Pattern Recognition}}}, 2017, pp. 2117--2125.

\bibitem{girshick2014rich}
R.~Girshick, J.~Donahue, T.~Darrell, and J.~Malik, ``Rich {{Feature
  Hierarchies}} for {{Accurate Object Detection}} and {{Semantic
  Segmentation}},'' in \emph{2014 {{IEEE Conference}} on {{Computer Vision}}
  and {{Pattern Recognition}} ({{CVPR}})}, Jun. 2014, pp. 580--587.

\bibitem{russakovsky2015imagenet}
O.~Russakovsky, J.~Deng, H.~Su, J.~Krause, S.~Satheesh, S.~Ma, Z.~Huang,
  A.~Karpathy, A.~Khosla, M.~Bernstein, A.~C. Berg, and L.~{Fei-Fei},
  ``{{ImageNet Large Scale Visual Recognition Challenge}},''
  \emph{International Journal of Computer Vision}, vol. 115, no.~3, pp.
  211--252, Dec. 2015.

\bibitem{maki2016enhanced}
J.~Maki, C.~McKinney, R.~Sellar, D.~Copley-Woods, D.~Gruel, D.~Nuding,
  M.~Valvo, T.~Goodsall, J.~McGuire, and T.~Litwin, ``Enhanced engineering
  cameras (eecams) for the mars 2020 rover,'' \emph{LPICo}, vol. 1980, p. 4132,
  2016.

\bibitem{cocoannotator}
J.~Brooks, ``{COCO Annotator},''
  \url{https://github.com/jsbroks/coco-annotator/}, 2019.

\bibitem{olson_2011}
E.~Olson, ``Apriltag: A robust and flexible visual fiducial system,''
  \emph{2011 IEEE International Conference on Robotics and Automation}, 2011.

\bibitem{everingham2010pascal}
M.~Everingham, L.~V. Gool, C.~K.~I. Williams, J.~Winn, and A.~Zisserman, ``The
  {{Pascal Visual Object Classes}} ({{VOC}}) {{Challenge}},''
  \emph{International Journal of Computer Vision}, vol.~88, no.~2, pp.
  303--338, Jun. 2010.

\bibitem{lin2014microsoft}
T.-Y. Lin, M.~Maire, S.~Belongie, J.~Hays, P.~Perona, D.~Ramanan,
  P.~Doll{\'a}r, and C.~L. Zitnick, ``Microsoft {{COCO}}: {{Common Objects}} in
  {{Context}},'' in \emph{Computer {{Vision}} \textendash{} {{ECCV}} 2014},
  ser. Lecture {{Notes}} in {{Computer Science}}, D.~Fleet, T.~Pajdla,
  B.~Schiele, and T.~Tuytelaars, Eds.\hskip 1em plus 0.5em minus 0.4em\relax
  {Springer International Publishing}, 2014, pp. 740--755.

\end{thebibliography}
